\documentclass[a4paper, 10pt]{article}

\usepackage{epsfig}
\usepackage{amsmath}
\usepackage{amssymb}
\usepackage[english,ruled,vlined]{algorithm2e}
\usepackage{verbatim}
\usepackage{subfigure}

\newcommand{\Bel}{\mathrm{Bel}}
\newcommand{\Pl}{\mathrm{Pl}}
\newcommand{\GPT}{\mathrm{GPT}}

\newcommand{\PCRmo}{\mathrm{PCR6}}

\newcommand{\conj}{\mathrm{c}}

\newcommand{\DS}{\mathrm{DS}}

\newtheorem{ffunctionmat}          {\quad Function}
\newenvironment{functionmat}       {\begin{ffunctionmat}\rm}{\end{ffunctionmat}}

\newcommand{\argmax}{\operatornamewithlimits{arg\,max}}

\title{Implementing general belief function framework with a practical codification for low complexity}
\author{Arnaud Martin \footnote{This work was carried out while the author was visiting DRDC (Defense Research and Development Canada) at Valcatier, Qu\'ebec, Canada, and is partially supported by the DGA (\emph{D\'el\'egation g\'en\'erale pour l'Armement}) and by BMO (\emph{Brest M\'etropole Oc\'eane}).} \\
ENSIETA E3I2-EA3876 \\
Arnaud.Martin@ensieta.fr}

\begin{document}
\maketitle

\begin{abstract}
In this chapter, we propose a new practical codification of the elements of the Venn diagram in order to easily manipulate the focal elements. In order to reduce the complexity, the eventual constraints must be integrated in the codification at the beginning. Hence, we only consider a reduced hyper power set $D_r^\Theta$ that can be $2^\Theta$ or $D^\Theta$. We describe all the steps of a general belief function framework. The step of decision is particularly studied, indeed, when we can decide on intersections of the singletons of the discernment space no actual decision functions are easily to use. Hence, two approaches are proposed, an extension of previous one and an approach based on the specificity of the elements on which to decide.  

The principal goal of this chapter is to provide practical codes of a general belief function framework for the researchers and users needing the belief function theory.

{\bf Keywords: DSmT, practical codification, DSmT decision, low complexity.}
\end{abstract}

\section{Introduction}
Today the belief function theory initiated by \cite{Dempster67, Shafer76} is recognized to propose one of the more complete theory for human reasoning under uncertainty, and have been applied in many kinds of applications \cite{Smets99}. This theory is based on the use of functions defined on the power set $2^\Theta$ (the set of all the subsets of $\Theta$), where $\Theta$ is the set of considered elements (called \textit{discernment space}), whereas the probabilities are defined only on $\Theta$. A \textit{mass function} or \textit{basic belief assignment}, $m$ is defined by the mapping of the power set $2^\Theta$ onto $[0,1]$ with:
\begin{equation}
\label{normDST}
\sum_{X\in 2^\Theta} m(X)=1.
\end{equation}
One element $X$ of $2^\Theta$, such as $m(X)>0$, is called \textit{focal} element. The set of focal elements for $m$ is noted ${\cal F}_m$. A mass function where $\Theta$ is a focal element, is called a \textit{non-dogmatic} mass functions. 

One of the main goal of this theory is the combination of information given by many experts. When this information can be written as a mass function, many combination rules can be used \cite{Martin07b}. The first combination rule proposed by Dempster and Shafer is the normalized conjunctive combination rule given for two basic belief assignments $m_1$ and $m_2$ and for all $X \in 2^\Theta$, $X\neq \emptyset$ by:
\begin{eqnarray}
m_\DS(X)=\displaystyle \frac{1}{1-k}\sum_{A\cap B =X} m_1(A)m_2(B),
\end{eqnarray}
where $k= \displaystyle \sum_{A\cap B =\emptyset} m_1(A)m_2(B)$ is the inconsistence of the combination.

However the high computational complexity, especially compared to the probability theory, remains a problem for more industrial uses. Of course, higher the cardinality of $\Theta$ is, higher the complexity becomes \cite{Wilson00}.  The combination rule of Dempster and Shafer is \#$P$-complete \cite{Orponen90}. Moreover, when combining with this combination rule, non-dogmatic mass functions, the number of focal elements can not decrease. 

Hence, we can distinguish two kinds of approaches to reduce the complexity of the belief function framework. First we can try to find optimal algorithms in order to code the belief functions and the combination rules based on M\"obius transform \cite{Kennes92,Smets02} or based on local computations \cite{Shenoy86} or to adapt the algorithms to particulars mass functions \cite{Shafer87,Barnett81}. Second we can try to reduce the number of focal elements by approximating the mass functions \cite{Voorbraak89,Tessem93,Bauer97,Denoeux01,Haenni02,Haenni03}, that could be particularly important for dynamic fusion.

In practical applications the mass functions contain at first only few focal elements \cite{Denoeux95,Appriou98}. Hence it seems interesting to only work with the focal elements and not with the entire space $2^\Theta$. That is not the case in all general developed algorithms \cite{Kennes92,Smets02}.

Now if we consider the extension of the belief function theory proposed by \cite{Dezert02}, the mass function are defined on the extension of the power set into the hyper power set $D^\Theta$ (that is the set of all the disjunctions and conjunctions of the elements of $\Theta$). This extension can be seen as a generalization of the classical approach (and it is also called DSmT for Dezert and Smarandache Theory \cite{Smarandache04,Smarandache06}). This extension is justified in some applications such as in \cite{Martin06c,Martin06a}. Try to generate $D^\Theta$ is not easy and becomes untractable for more than 6 elements in $\Theta$ \cite{Dezert04c}.

In \cite{Dezert04d}, a first proposition have been proposed to order elements of hyper power set for matrix calculus such as \cite{Kennes92,Smets02} made in $2^\Theta$. But as we said herein, in real applications it is better to only manipulate the focal elements. Hence, some authors propose algorithms considering only the focal elements \cite{Denoeux01,Djiknavorian06,Martin06b}. In the previous volume \cite{Smarandache06}, \cite{Djiknavorian06} have proposed Matlab\footnote{Matlab is a trademark of The MathWorks, Inc.} codes for DSmT hybrid rule. These codes are a preliminary work, but first it is really not optimized for Matlab and second have been developed for a dynamic fusion.

Matlab is certainly not the best program language to reduce the speed of processing, however most of people using belief functions do it with Matlab. 

In this chapter, we propose a codification of the focal elements based on a codification of $\Theta$ in order to program easily in Matlab a general belief function framework working for belief functions defined on $2^\Theta$ but also on $D^\Theta$.

Hence, in the following section we recall a short background of belief function theory. In section~\ref{codification} we introduce our practical codification for a general belief function framework. In this section, we describe all the steps to fuse basic belief assignments in the order of necessity: the codification of $\Theta$, the addition of the constraints, the codification of focal elements, the step of combination, the step of decision, if necessary the generation of a new power set: the \textit{reduced hyper power set} $D_r^\Theta$ and for the display, the decoding. We particularly investigate the step of the decision for the DSmT. In section~\ref{codes} we give the major part of the Matlab codes of this framework.

\section{Short background of belief functions theory}
\label{background}
In the DSmT, the mass functions $m$ are defined by the mapping of the hyper power set $D^\Theta$ onto $[0,1]$ with:
\begin{equation}
\label{normDTheta}
\sum_{X\in D^\Theta} m(X)=1,
\end{equation}
with less terms in the sum than in the equation~\eqref{normDTheta}.

In the more general model, we can add constraints on some elements of $D^\Theta$, that means that some elements can never be focal elements. Hence, if we add the constraints that all the intersections of elements of $\Theta$ are impossible (\emph{i.e.} empty) we recover $2^\Theta$. So, the constraints given by the application can drastically reduce the number of possible focal elements and so the complexity of the framework. On the contrary of the suggestion given by the flowchart on the cover of the book \cite{Smarandache04} and the proposed codes in \cite{Djiknavorian06}, we think that the constraints must be integrated directly in the codification of the focal elements of the mass functions as we shown in section~\ref{codification}. Hereunder, the hyper power set $D^\Theta$ taking into account the constraints is called the \textit{reduced hyper power set} and noted $D_r^\Theta$. Hence, $D_r^\Theta$ can be $D^\Theta$, $2^\Theta$, have a cardinality between these two power sets or inferior to these two power sets. So the normality condition is given by:
\begin{equation}
\label{normalisation}
\sum_{X\in D_r^\Theta} m(X)=1.
\end{equation}

Once defined the mass functions coming from numerous sources, many combination rules are possible (see \cite{Daniel06,Smarandache06b,Martin06c,Smets06,Martin07b} for recent reviews of the combination rules). The most of the combination rules are based on the conjunctive combination rule, given for mass functions defined on $2^\Theta$ by:
\begin{eqnarray}
\label{conjunctive}
m_\conj(X)=\sum_{Y_1 \cap ... \cap Y_s = X} \prod_{j=1}^s m_j(Y_j),
\end{eqnarray} 
where $Y_j \in 2^\Theta$ is the response of the source $j$, and $m_j(Y_j)$ the corresponding basic belief assignment. This rule is commutative, associative, not idempotent, and the major problem that try to resolve the majority of the rules is the increasing of the belief on the empty set with the number of sources and the cardinality of $\Theta$ \cite{Martin08a}. Now, in $D^\Theta$ without any constraint, there is no empty set, and the conjunctive rule given by the equation~\eqref{conjunctive} for all $X \in D^\Theta$ with $Y_j \in D_r^\Theta$ can be used. If we have some constraints, we must to transfer the belief $m_\conj(\emptyset)$ on other elements of the reduced hyper power set. There is no optimal combination rule, and we cannot achieve this optimality for general applications.  

The last step in a general framework for information fusion system is the decision step. The decision is also a difficult task because no measures are able to provide the best decision in all the cases. Generally, we consider the maximum of one of the three functions: credibility, plausibility, and pignistic probability. Note that other decision functions have been proposed \cite{Dezert08}.

In the context of the DSmT the corresponding generalized functions have been proposed \cite{Dezert04a, Smarandache04}.
The generalized credibility $\Bel$ is defined by:
\begin{eqnarray}
\label{bel}
\Bel(X)=\sum_{Y \in D_r^\Theta, Y\subseteq X, Y \not\equiv \emptyset} m(Y)
\end{eqnarray}
The generalized plausibility $\Pl$ is defined by:
\begin{eqnarray}
\label{pl}
\Pl(X)=\sum_{Y \in D_r^\Theta, X \cap Y\not\equiv \emptyset} m(Y)
\end{eqnarray}
The generalized pignistic probability is given for all $X \in D_r^\Theta$, with $X \neq \emptyset$ is defined by:
\begin{eqnarray}
\label{betp}
\GPT(X)=\sum_{Y \in D_r^\Theta, Y \not\equiv \emptyset} \frac{{\cal C_M}(X \cap Y)}{{\cal C_M}(Y)} m(Y),
\end{eqnarray}
where ${\cal C_M}(X)$ is the DSm cardinality corresponding to the number of parts of $X$ in the Venn diagram of the problem \cite{Dezert04a, Smarandache04}. Generally in $2^\Theta$, the maximum of these functions is taken on the elements in $\Theta$. In this case, with the goal to reduce the complexity we only have to calculate these functions on the singletons. However, first, there exist methods providing decision on $2^\Theta$ such as in \cite{Appriou05} and that can be interesting in some application \cite{Martin08b}, and secondly, the singletons are not the more precise elements on $D_r^\Theta$. Hence, to calculate these functions on the entire reduced hyper power set could be necessary, but the complexity could not be inferior to the complexity of $D_r^\Theta$ and that can be a real problem if there are few constraints.

\section{A general belief function framework}
\label{codification}
We introduce here a practical codification in order to consider all the previous remarks to reduce the complexity:
\begin{itemize}
\item only manipulate focal elements,
\item add constraints on the focal elements before combination, and so work on $D_r^\Theta$,
\item a codification easy for union and intersection operations with programs such as Matlab.
\end{itemize}

We first give the simple idea of the practical codification for enumerating the distinct parts of the Venn diagram and so a codification of the discernment space $\Theta$. Then we explain how simply add the constraints on the distinct elements of $\Theta$ and so the codification of the focal elements. The subsections~\ref{comb} and \ref{dec} show how to combine and decide with this practical codification, giving a particular reflexion on the decision in DSmT. The subsection ~\ref{genDThetar} presents the generation of $D_r^\Theta$ and the subsection~\ref{decoding} the decoding.

\subsection{A practical codification}
\begin{figure}[htb]
  \begin{center}
      \includegraphics[height=6cm]{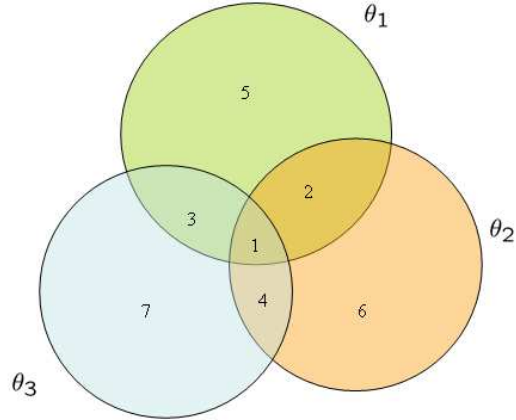}
  \end{center}
  \caption{Codification for $\Theta=\{\theta_1,\theta_2,\theta_3\}$.}
  \label{codificationTheta3}
\end{figure}

\begin{figure}[htb]
  \begin{center}
      \includegraphics[height=8cm]{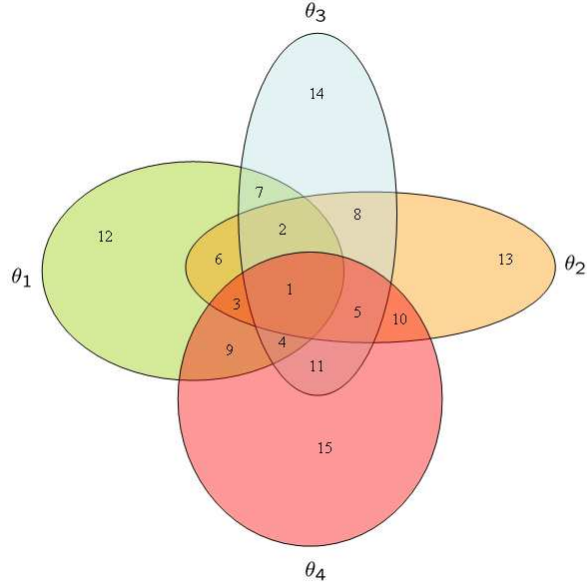}
  \end{center}
  \caption{Codification for $\Theta=\{\theta_1,\theta_2,\theta_3,\theta_4\}$.}
  \label{codificationTheta4}
\end{figure}

The simple idea of the practical codification is based on the affectation of an integer number in $[1;2^n-1]$ to each distinct part of the Venn diagram that contains $2^n-1$ distinct parts with $n=|\Theta|$. The figures~\ref{codificationTheta3} and \ref{codificationTheta4} illustrate the codification for respectively $\Theta=\{\theta_1,\theta_2,\theta_3\}$ and $\Theta=\{\theta_1,\theta_2,\theta_3,\theta_4\}$ with the code given in section~\ref{codes}. Of course other repartitions of these integers are possible. 

Hence, for example the element $\theta_1$ is given by the concatenation of 1, 2, 3 and 5 for $|\Theta|=3$ and by the concatenation of 1, 2, 3, 4, 6, 7, 9 and 12 for  $|\Theta|=4$. We will note respectively $\theta_1=[1~ 2~ 3~ 5]$ and $\theta_1=[1~ 2~ 3~ 4~ 6~ 7~ 9~ 12]$ for $|\Theta|=3$ and for $|\Theta|=4$, with increasing order of the integers. Hence, $\Theta$ is given respectively for $|\Theta|=3$ and $|\Theta|=4$ by:
$$\Theta=\{[1~ 2~ 3~ 5], [1~ 2~ 4~ 6], [1~ 3~ 4~ 7]\}$$
and
$$\Theta=\{[1~ 2~ 3~ 4~ 6~ 7~ 9~ 12], [1~ 2~ 3~ 5~ 6~ 8~ 10~ 13], [1~ 2~ 4~ 5~ 7~ 8~ 11~ 14], [1~ 3~ 4~ 5~ 9~ 10~ 11~ 15]\}.$$
The number of integers for the codification of one element $\theta_i \in \Theta$ is given by:
\begin{eqnarray}
\label{number_sing}
1+\sum_{i=1}^{n-1}C_{n-1}^i,
\end{eqnarray}
with $n=|\Theta|$ and $C_n^p$ the number of $p$-uplets with $n$ numbers. The number 1 will be still by convention the intersection of all the elements of $\Theta$. The codification of $\theta_1 \cap \theta_3$ is given by $[1~ 3]$ for $|\Theta|=3$ and $[1~ 2~ 4~ 7]$ for $|\Theta|=4$. And the codification of $\theta_1 \cup \theta_3$ is given by $[1~ 2~ 3~ 4~ 5~ 7]$ for $|\Theta|=3$ and $[1~ 2~ 3~ 4~ 6~ 7~ 9~ 12]$ for $|\Theta|=4$. 

In order to reduce the complexity, especially using more hardware language than Matlab, we could use binary numbers instead of the integer numbers. 

\begin{figure}[htb]
  \begin{center}
      \includegraphics[height=6cm]{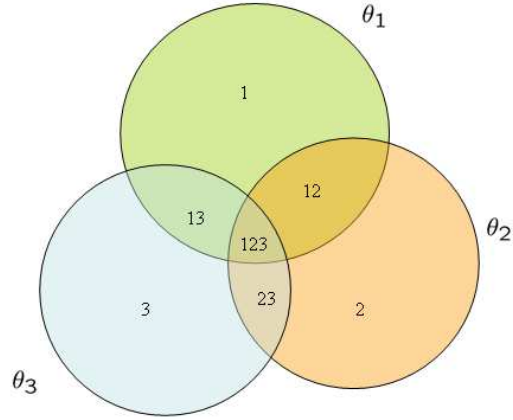}
  \end{center}
  \caption{Smarandache's codification for $\Theta=\{\theta_1,\theta_2,\theta_3\}$.}
  \label{SmaCodificationTheta3}
\end{figure}

The Smarandache's codification \cite{Dezert04c}, was introduce for the enumeration of distinct parts of a Venn diagram. If $|\Theta|=n$, $<i>$ denotes the part of $\theta_i$ with no covering with other $\theta_j$, $i\neq j$. $<ij>$ denotes the part of $\theta_i \cap \theta_j$ with no covering with other parts of the Venn diagram. So if $n=2$, $\theta_1 \cap \theta_2=\{<12>\}$ and if $n=3$, $\theta_1 \cap \theta_2=\{<12>,<123>\}$, see the figure~\ref{SmaCodificationTheta3} for an illustration for $n=3$. The authors note a problem for $n\geq 10$, but if we introduce space in the codification we can conserve integers instead of other symbols and we write $<1~2~3>$ instead of $<123>$. 

On the contrary of the Smarandache's codification, the proposed codification gives only one integer number to each part of the Venn diagram. This codification is more complex for the reader then the Smarandache's codification. Indeed, the reader can understand directly the Smarandache's codification thanks to the mining of the numbers knowing the $n$: each disjoint part of the Venn diagram is seen as an intersection of the elements of $\Theta$. More exactly, this is a part of the intersections. For example, $\theta_1\cap \theta_2$ is given with  the Smarandache's codification by $\{<12>\}$ if $n=2$ and by  $\{<12>,<123>\}$ if $n=3$. With the codification practical codification the same element has also different codification according to the number $n$. For the previous example $\theta_1\cap \theta_2$ is given by $[1]$ if $n=2$, and by $[1~2]$ if $n=3$. 

The proposed codification is more practical for computing union and intersection operations and the DSm cardinality, because only one integer represent one of the distinct parts of the Venn diagram. With the Smarandache's codification computing union and intersection operations and the DSm cardinality could be very similar than with the practical codification, but adding a routine in order to treat the code of one part of the Venn diagram.

Hence, we propose to use the proposed codification to compute union, intersection and DSm cardinality, and the Smarandache's codification, easier to read, to present the results in order to safe eventually a scan of $D_r^\Theta$.

\subsection{Adding constraints}
With this codification, adding constraints is very simple and can reduce rapidly the number of integers. \emph{E.g.} assume that in a given application we know $\theta_1 \cap \theta_3 \equiv \emptyset$ (\emph{i.e.} $\theta_1 \cap \theta_3 \notin D_r^\Theta$), that means that the integers $[1~ 3]$ for $|\Theta|=3$ and $[1~ 2~ 4~ 7]$ for $|\Theta|=4$ do not exist $\Theta$. Hence, the codification of $\Theta$ with the reduced discernment space, noted $\Theta_r$, is given respectively for $|\Theta|=3$ and $|\Theta|=4$ by:
$$\Theta_r=\{[2~ 5], [2~ 4~ 6], [4~ 7]\}$$
and
$$\Theta_r=\{[3~ 6~ 9~ 12], [3~ 5~ 6~ 8~ 10~ 13], [5~ 8~ 11~ 14], [3~ 5~ 9~ 10~ 11~ 15]\}.$$
Generally we have $|\Theta|=|\Theta_r|$, but it is not necessary if a constraint gives $\theta_i \equiv \emptyset$, with $\theta_i \in \Theta$. This can happen in dynamic fusion, if one element of the discernment space can disappear.  

Thereby, the introduction of the simple constraint $\theta_1 \cap \theta_3 \equiv \emptyset$ in $\Theta$, includes all the other constraints that follow from it such as the intersection of all the elements of $\Theta$ is empty. In \cite{Djiknavorian06} all the constraints must be given by the user.

\subsection{Codification of the focal elements}

In $D_r^\Theta$, the codification of the focal elements is given from the reduced discernment space $\Theta_r$. The codification of an union of two elements of $\Theta$ is given by the concatenation of the codification of the two elements using $\Theta_r$. The codification of an intersection of two elements of $\Theta$ is given by the common numbers of the codification of the two elements using $\Theta_r$. In the same way, the codification of an union of two focal elements is given by the concatenation of the codification of the two focal elements and the codification of an intersection of two focal elements is given by the common numbers of the codification of the two focal elements. In fact, for union and intersection operations we only consider one element as the set of the numbers given in its codification.

Hence, with the previous example (we assume $\theta_1 \cap \theta_3 \equiv \emptyset$, with $|\Theta|=3$ or $|\Theta|=4$), if the following elements $\theta_1 \cap \theta_2$, $\theta_1 \cup \theta_2$ and $(\theta_1 \cap \theta_2) \cup \theta_3$ are some focal elements, there are coded for $|\Theta|=3$ by:
$$\theta_1 \cap \theta_2=[2],$$
$$\theta_1 \cup \theta_2 = [2~4~5~6],$$
$$(\theta_1 \cap \theta_2) \cup \theta_3=[2~4~7],$$
and for $|\Theta|=4$ by:
$$\theta_1 \cap \theta_2=[3~6],$$
$$\theta_1 \cup \theta_2 = [3~ 5~ 6~ 8~ 9~ 10~12~ 13],$$
$$(\theta_1 \cap \theta_2) \cup \theta_3=[3~5~6~ 8~ 11~ 14].$$

%\subsubsection{Remarks and advantages}
The DSm cardinality ${\cal C_M}(X)$ of one focal element $X$ is simply given by the number of integers in the codification of $X$. The DSm cardinality of one singleton is given by the equation~\eqref{number_sing}, only if there is none constraint on the singleton, and inferior otherwise. 

The previous example with the focal element $(\theta_1 \cap \theta_2) \cup \theta_3$ illustrates well the easiness to deal with the brackets in one expression. The codification of the focal elements can be made with any brackets.

\subsection{Combination}
\label{comb}
In order to manage only the focal elements and their associated basic belief assignment, we can use a list structure \cite{Denoeux01,Djiknavorian06,Martin06b}. The intersection and union operations between two focal elements coming from two mass functions are made as described before. If the intersections between two focal elements is empty the associated codification is $[~]$. Hence the conjunctive combination rule algorithm can be done by the algorithm~\ref{AlgConj}. The disjunctive combination rule algorithm is exactly the same by changing $\cap$ in $\cup$.

\begin{algorithm}
  \KwData{$n$ experts $ex$: $ex[1] \ldots ex[n]$, $ex[i].focal$, $ex[i].bba$}
  \KwResult{Fusion of $ex$ by conjunctive rule: $conj$}
  %\For{$i$ = 1 to $n$}{
%    	$nbfocal[i]$ $\gets$ size of $ex[i].focal$\;}
    	
  $extmp$ $\gets$ $ex[1]$\;
  
  \For{$e$ = 2 to $n$}{
  $comb \gets \emptyset$\;
  
  \ForEach{$foc1$ in $extmp.focal$}{
  	\ForEach{$foc2$ in $ex[e].focal$}{
  		$tmp \gets extmp.focal(foc1) \cap ex[e].focal(foc2)$\;
  		$comb.focal \gets tmp$\;
  		$comb.bba \gets extmp.bba(foc1) \times ex[e].bba(foc2)$\;	
  	 }
  	}
  	Concatenate same focal in $comb$\;
  	$extmp \gets comb$\;
  	}
  $conj \gets extmp$\;

  \caption{Conjunctive rule}
  \label{AlgConj}
\end{algorithm}

Once again, the interest of the codification is for the intersection and union operations. Hence in Matlab, we do not need to redefine these operations as in \cite{Djiknavorian06}.

For more complicated combination rules such as $\PCRmo$, we have generally to conserve the intermediate calculus in order to transfer the partial conflict. Algorithms for these rules have been proposed in \cite{Martin06b}, and Matlab codes are given in section~\ref{codes}.

\subsection{Decision}
\label{dec}
\begin{figure}[htb]
  \begin{center}
      \includegraphics[height=7cm]{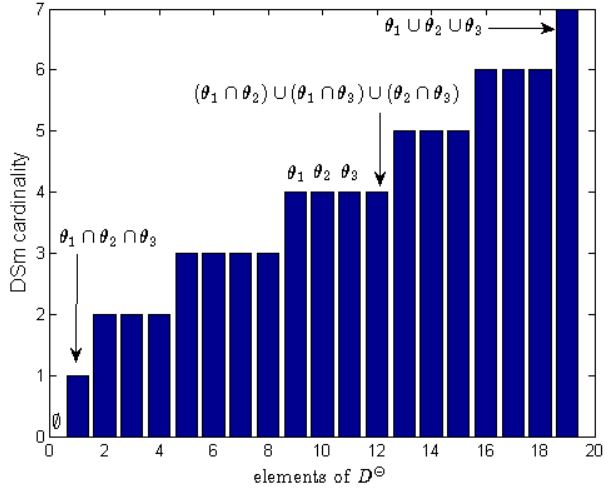}
  \end{center}
  \caption{DSm cardinality ${\cal C_M}(X)$, $\forall X \in D^\Theta$ with $|\Theta|=3$.}
  \label{cardinalityDTheta3}
\end{figure}

\begin{figure}[htb]
  \begin{center}
      \includegraphics[height=7cm]{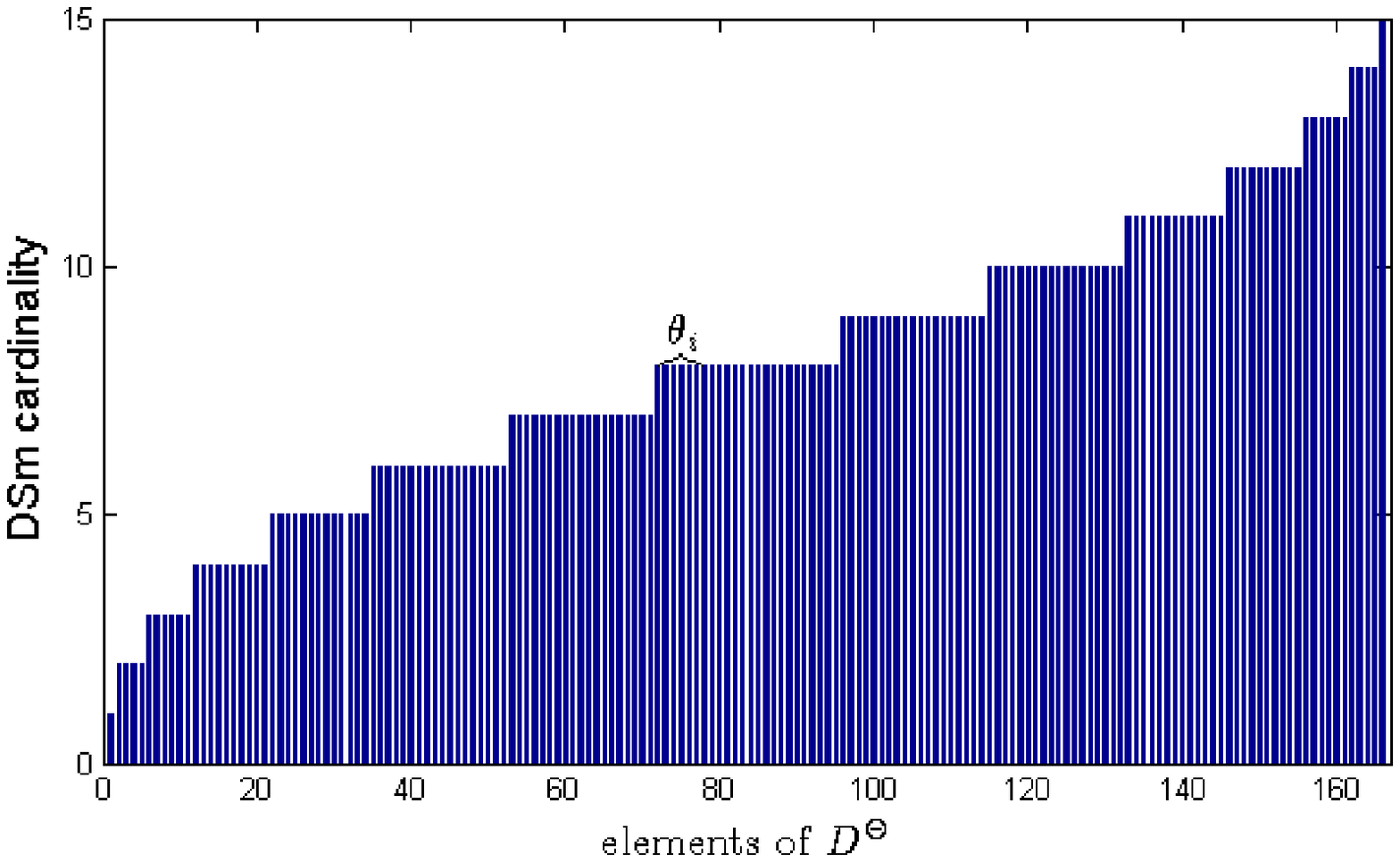}
  \end{center}
  \caption{DSm cardinality ${\cal C_M}(X)$, $\forall X \in D^\Theta$ with $|\Theta|=4$.}
  \label{cardinalityDTheta4}
\end{figure}

As we write before, we can decide with one of the functions given by the equations~\eqref{bel}, \eqref{pl}, or \eqref{betp}. These functions are increasing functions. Hence generally in $2^\Theta$, the decision is taken on the elements in $\Theta$ by the maximum of these functions. In this case, with the goal to reduce the complexity, we only have to calculate these functions on the singletons. However, first, we can provide a decision on any element of $2^\Theta$ such as in \cite{Appriou05} that can be interesting in some applications \cite{Martin08b}, and second, the singletons are not the more precise or interesting elements on $D_r^\Theta$. The figures~\ref{cardinalityDTheta3} and \ref{cardinalityDTheta4} show the DSm cardinality ${\cal C_M}(X)$, $\forall X \in D^\Theta$ with respectively $|\Theta|=3$ and $|\Theta|=4$. The specificity of the singletons (given by the DSm cardinality) appears at a central position in the set of the specificities of the elements in $D^\Theta$. 

Hence, to calculate these decision functions on all the reduced hyper power set could be necessary, but the complexity could not be inferior to the complexity of $D_r^\Theta$ and that can be a real problem. The more reasonable approach is to consider either only the focal elements or a subset of $D_r^\Theta$ on which we calculate decision functions.

\subsubsection{Extended weighted approach}
Generally in $2^\Theta$, the decisions are only made on the singletons \cite{Denoeux97,Smets05}, and only few approaches propose a decision on $2^\Theta$. In order to provide decision on any elements of $D_r^\Theta$, we can first extend the principle of the proposed approach in \cite{Appriou05} on $D_r^\Theta$. This approach is based on the weighting of the plausibility with a Bayesian mass function taking into account the cardinality of the elements of $2^\Theta$. 

In a general case, if there is no constraint, the plausibility is not interesting because all elements contain the intersection of all the singletons of $\Theta$. According the constraints the plausibility could be applied. 

Hence, we generalize here the weighted approach to $D_r^\Theta$ for every decision function $f_d$ (plausibility, credibility, pignistic probability, ...). We note $f_{wd}$ the weighted decision function given for all $X\in D_r^\Theta$ by:
\begin{eqnarray}
\label{f_wd}
	f_{wd}(X)= m_d(X)f_d(X),
\end{eqnarray}
where $m_d$ is a basic belief assignment given by:
\begin{eqnarray}
m_d(X)=K_d \lambda_X \left(\frac{1}{{\cal C_M}(X)^s}\right),
\end{eqnarray}
$s$ is a parameter in $[0,1]$ allowing a decision from the intersection of all the singletons ($s=1$) (instead of the singletons in $2^\Theta$) until the total indecision $\Theta$ ($s=0$). $\lambda_X$ allows the integration of the lack of knowledge on one of the elements $X$ in $D_r^\Theta$. The constant $K_d$ is the normalization factor giving by the condition of the equation~\eqref{normalisation}. Thus we decide the element $A$:
\begin{eqnarray}
\label{DecAppriou}
	A=\argmax_{X \in D_r^\Theta}f_{wd}(X),
\end{eqnarray}

If we only want to decide on whichever focal element of $D_r^\Theta$, we only consider $X\in {\cal F}_m$ and we decide:
\begin{eqnarray}
	A=\argmax_{X \in {\cal F}_m} f_{wd}(X),
\end{eqnarray}
with $f_{wd}$ given by the equation~\eqref{f_wd} and:
\begin{eqnarray}
m_d(X)=K_d \lambda_X \left(\frac{1}{{\cal C_M}(X)^s}\right), \, \forall X\in {\cal F}_m,
\end{eqnarray}
$s$ and $K_d$ are both parameters defined above. 

\subsubsection{Decision according to the specificity}

\begin{figure}[htb]
  \begin{center}
     \includegraphics[height=7cm]{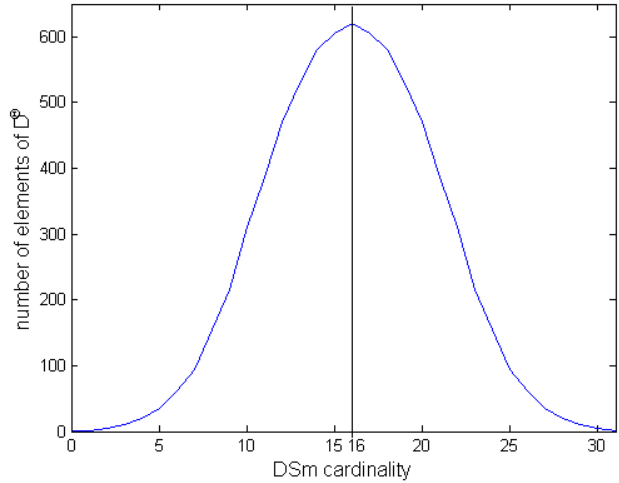}
  \end{center}
  \caption{Number of elements of $D^\Theta$ for $|\Theta|=5$, with the same DSm cardinality.}
  \label{distributionCardinalityDTheta5}
\end{figure}
The cardinality ${\cal C_M}(X)$ can be seen as a specificity measure of $X$. The figures~\ref{cardinalityDTheta3} and \ref{cardinalityDTheta4} show that for a given specificity there is different kind of elements such as singletons, unions of intersections or intersections of unions. The figure~\ref{distributionCardinalityDTheta5} shows well the central role of the singletons (the DSm cardinality of the singletons for $|\Theta|$=5 is 16), but also that there is many other elements (619) with exactly the same cardinality. Hence, it could be interesting to precise the specificity of the elements on which we want to decide. This is the role of $s$ in the Appriou approach. Here we propose to directly give the wanted specificity or an interval of the wanted specificity in order to build the subset of $D_r^\Theta$ on which we calculate decision functions. Thus we decide the element $A$:
\begin{eqnarray}
	A=\argmax_{X \in {\cal S}} f_d(X),
\end{eqnarray}
where $f_d$ is the chosen decision function (credibility, plausibility, pignistic probability, ...) and 
\begin{eqnarray}
{\cal S}=\left\{X\in D_r^\Theta ; min_S \leq {\cal C_M}(X) \leq max_S  \right\},
\end{eqnarray}
with $min_S$ and $max_S$ respectively the minimum and maximum of the specificity of the wanted elements. If $min_S \neq max_S$, if have to chose a pondered decision function for $f_d$ such as $f_{wd}$ given by the equation~\eqref{f_wd}. 

However, in order to find all $X\in{\cal S}$ we must scan $D_r^\Theta$. To avoid to scan all $D_r^\Theta$, we have to find the cardinality of ${\cal S}$, but we can only calculate an upper bound of the cardinality, unfortunately never reached. Let define the number of elements of the Venn diagram $n_V$. This number is given by:
\begin{eqnarray}
	n_V={\cal C_M}\left(\displaystyle \bigcup_{i=1}^n \theta_i\right),
\end{eqnarray}
where $n$ is the cardinality of $\Theta_r$ and $\theta_i \in \Theta_r$. Recall that the DSm cardinality is simply given by the number of integers of the codification. The upper bound of the cardinality of ${\cal S}$ is given by:
\begin{eqnarray}
	|{\cal S}|<\sum_{s=min_S}^{max_S}C_{n_V}^s,
\end{eqnarray}
where $C_{n_V}^s$ is the number of combinations of $s$ elements among $n_V$. Note that it also works if $min_S=0$ for the empty set.

\subsection{Generation of $D_r^\Theta$}
\label{genDThetar}

\begin{figure}
\leavevmode 
\begin{tabular}{cc}  
%\hspace*{-1.7cm}
\!\!\!\!\!\!\!\! \subfigure[\label{cardinalityDThetaRed4L}Elements are ordered by increasing DSm cardinality.]{\scalebox{0.38}{\includegraphics{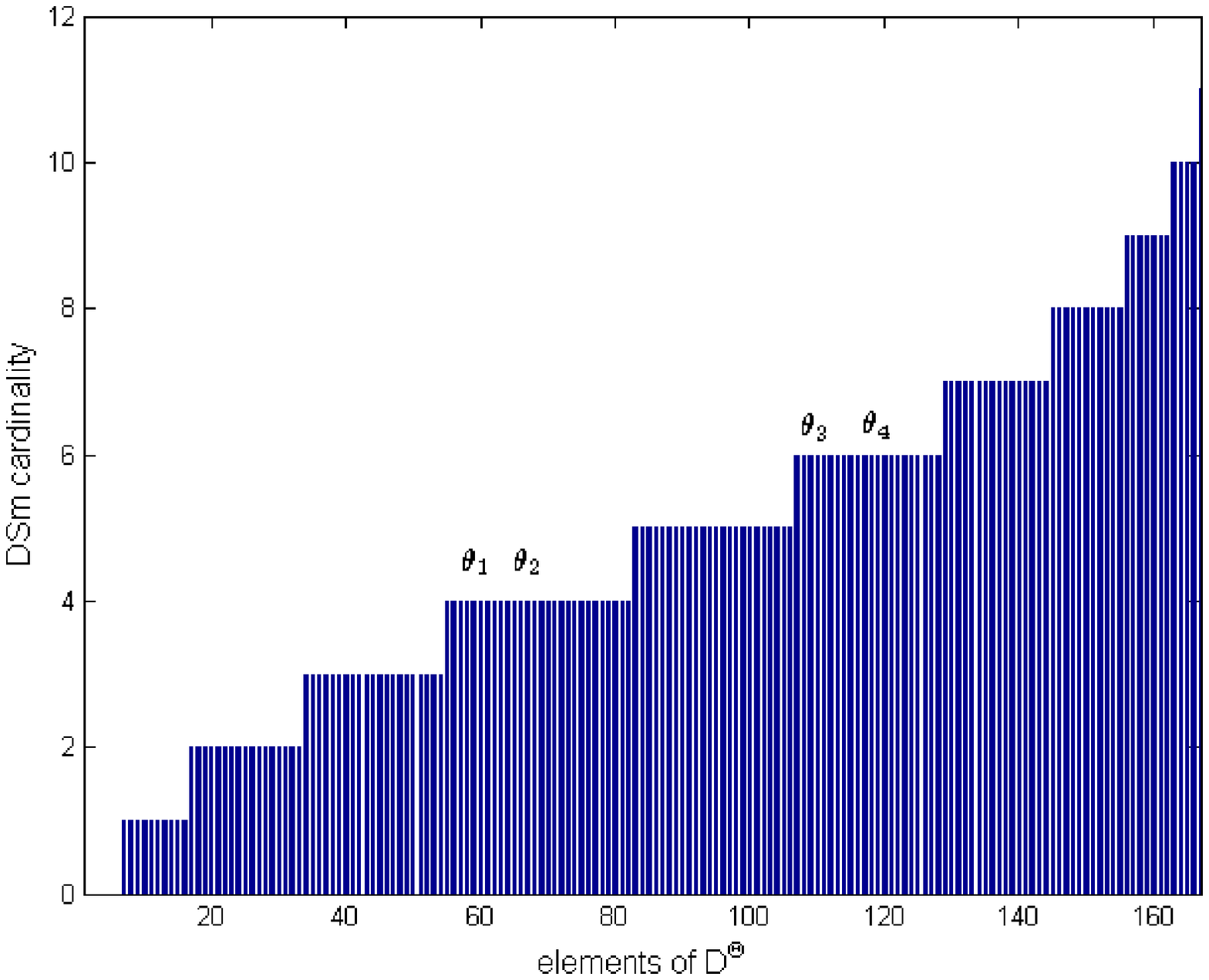}}}
%\hspace*{-1.7cm}& 
%\hspace*{-1.5cm}
\subfigure[\label{cardinalityDThetaRed4R}Elements are ordered with the same order than the figure~\ref{cardinalityDTheta4}.]{\scalebox{0.38}{\includegraphics{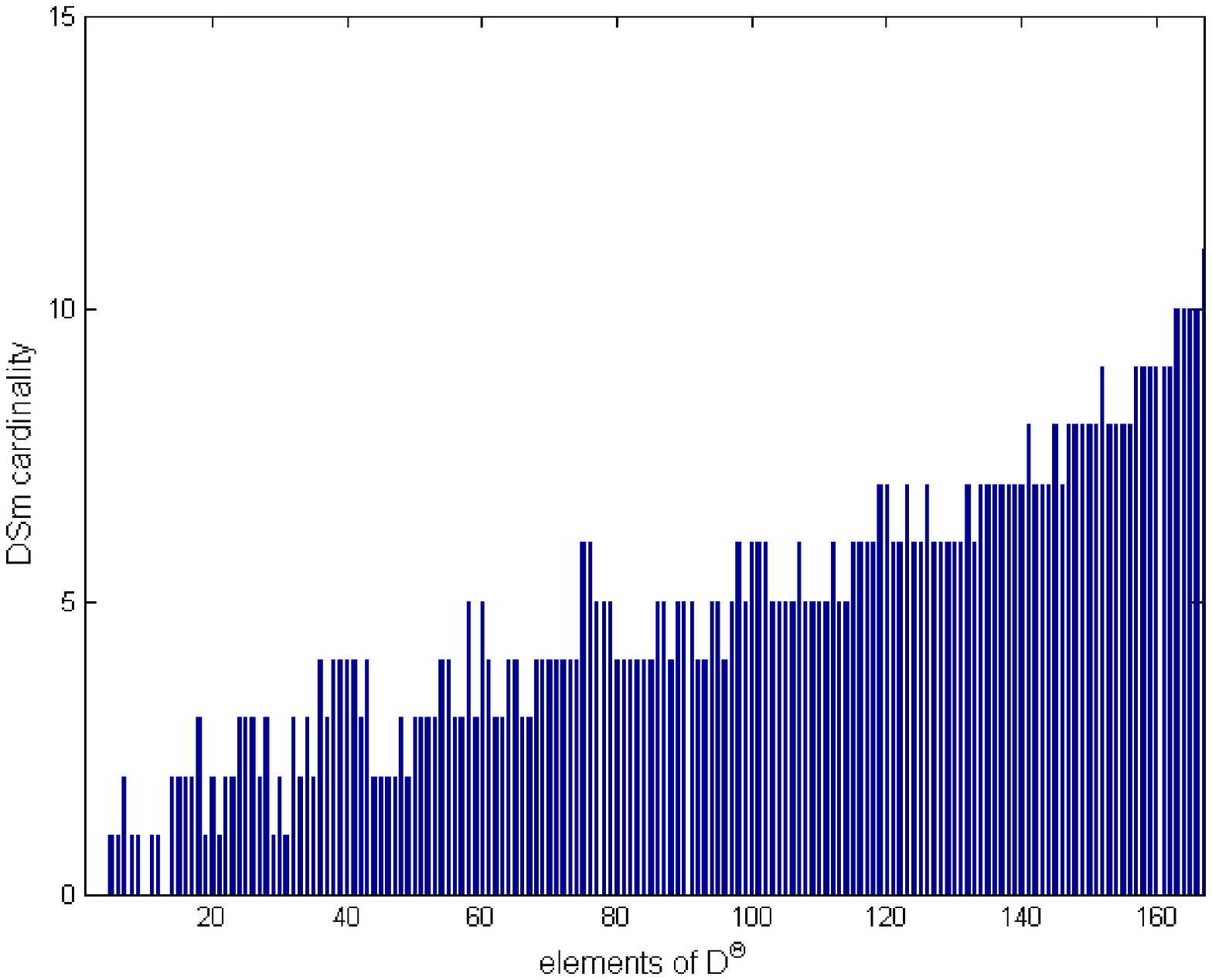}}}
\end{tabular}
\caption{DSm cardinality ${\cal C_M}(X)$, $\forall X \in D_r^\Theta$ with $|\Theta|=4$ and $\theta_1 \cap \theta_2 \equiv \emptyset$.}
\label{cardinalityDThetaRed4} 
\end{figure}

\begin{figure}[htb]
  \begin{center}
     \includegraphics[height=7cm]{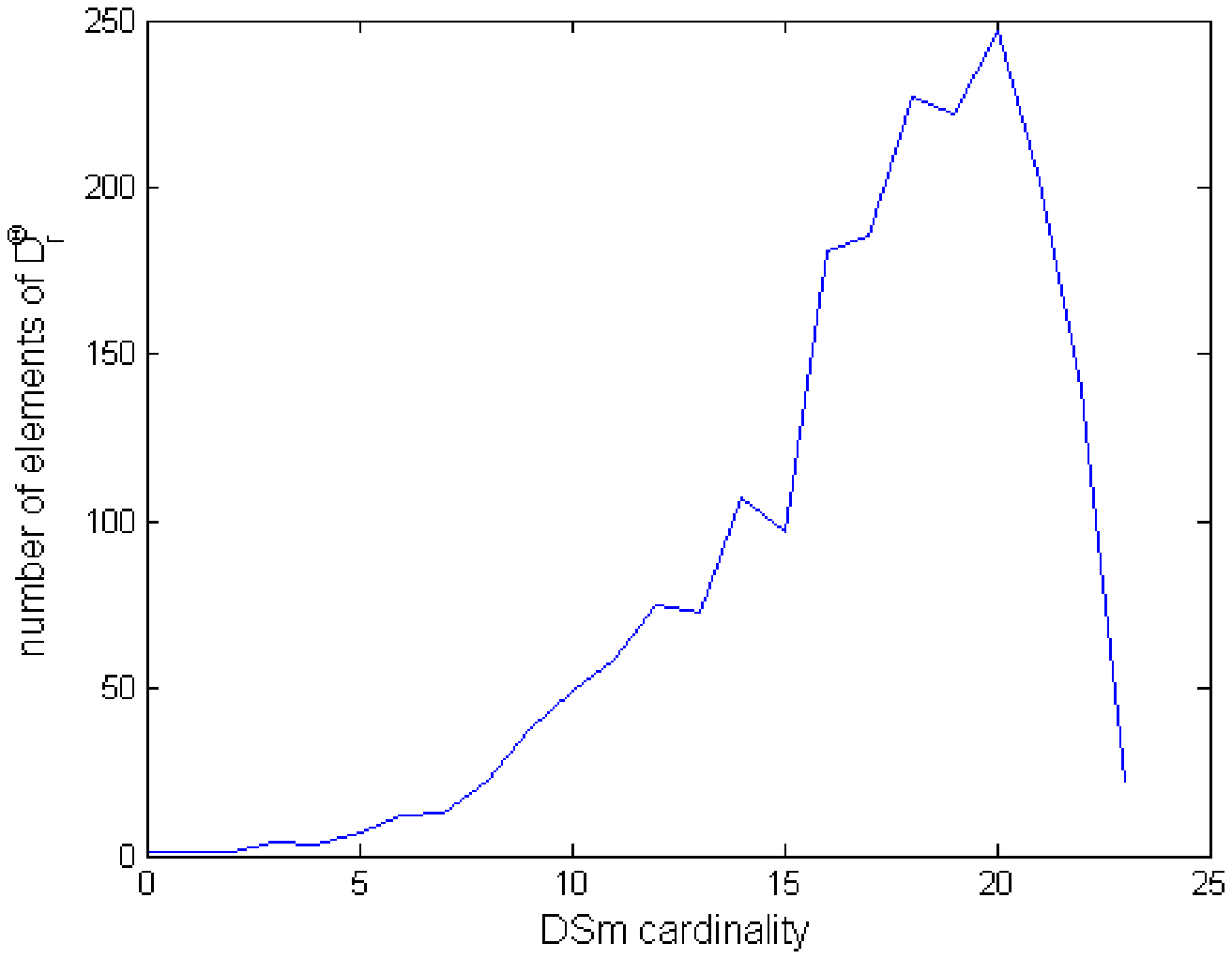}
  \end{center}
  \caption{Number of elements of $D_r^\Theta$ for $|\Theta|=5$ and $\theta_1 \cap \theta_2 \equiv \emptyset$, with the same positive DSm cardinality.}
  \label{distributionCardinalityDThetaRed5}
\end{figure}

The generation of $D_r^\Theta$ could have the same complexity than the generation of $D^\Theta$ if there is none constraint given by the user. Today, the complexity of the generation of $D^\Theta$ is the complexity of the proposed code in \cite{Dezert04c}. Assume for example, the simple constrain $\theta_1 \cap \theta_2 \equiv \emptyset$. First, the figures~\ref{cardinalityDThetaRed4L} 	and \ref{cardinalityDThetaRed4R} show the DSm cardinality for the elements of $D_r^\Theta$ with $|\Theta|=4$ and the previous given constraint. On the left figure, the elements are ordered by increasing DSm cardinality and on the right figure with the same order than the figure~\ref{cardinalityDTheta4}. We can observe that the cardinality of the elements have naturally decreased and the number of non empty elements also. This is more interesting if the cardinality of $\Theta$ is higher. Figure~\ref{distributionCardinalityDThetaRed5} presents for a given positive DSm cardinality, the number of elements of $D_r^\Theta$ for $|\Theta|=5$ and with the same constraint $\theta_1 \cap \theta_2 \equiv \emptyset$. Compared to the figure~\ref{distributionCardinalityDTheta5}, the total number of non empty elements (the integral of the curve) is considerably lower.

%\begin{figure}[h]
% \begin{minipage}{0.45\textwidth}
% 	\includegraphics[height=7cm]{CardinalityDThetaRed4_left.png}
% 		\caption{Elements are ordered by increasing DSm cardinality.}
% 		\label{cardinalityDThetaRed4L}
% 		\end{minipage}
% 		\hspace{0.1\textwidth}
% \begin{minipage}{0.45\textwidth}
% 	\includegraphics[height=7cm]{CardinalityDThetaRed4_right.png}
% 	\caption{Elements are ordered with the same order than the figure~\ref{cardinalityDTheta4}.}
% 		\label{cardinalityDThetaRed4R}
% 		\end{minipage}
%  \caption{DSm cardinality ${\cal C_M}(X)$, $\forall X \in D_r^\Theta$ with $|\Theta|=4$ and $\theta_1 \cap \theta_2 \equiv \emptyset$.} 
%  \label{cardinalityDThetaRed4R}
%\end{figure} 

Thus, we have to generate $D_r^\Theta$ and not $D^\Theta$. The generation of $D^\Theta$ (see \cite{Dezert04c} for more details) is based on the generation of monotone boolean functions. A monotone boolean function $f_{mb}$ is a mapping of $(x_1,...,x_b)\in\{0,1\}^b$ to a single binary output such as $\forall {\bf x},{\bf x}' \in \{0,1\}^b$, with ${\bf x}\preccurlyeq {\bf x'}$ then $f_{mb}({\bf x})\leq f_{mb}({\bf x'})$. Hence, a monotone boolean function is defined by the values of the $2^b$ elements $(x_1,...,x_b)$, and there is $|D^b|$ different monotone boolean functions. All the values of all these monotone boolean function can be represented by a $|D^b|\times 2^b$ matrix. If we multiply this matrix by the vector of all the possible intersections of the singletons in $\Theta$ with $|\Theta|=b$ (there is $2^b$ intersections) given an union of intersections, we obtain all the elements of $D^\Theta$. We can also use the basis of all the unions of $\Theta$ (and obtain the intersections of unions), but with our codification the unions are coded with more integer numbers. So, the intersection basis is preferable. 

Moreover, if we have some constraints (such as $\theta_1 \cap \theta_2\equiv \emptyset$), some elements of the intersection basis can be empty. So we only need to generate a $|D^b|\times n_b$ matrix where $n_b$ is the number of non empty intersections of elements in $\Theta_r$. For example, with the constraint given in example for $|\Theta|=3$, the basis is given by: $\emptyset$, $\theta_1$, $\theta_2$, $\theta_3$, $\theta_1 \cap \theta_3$, $\theta_2 \cap \theta_3$, and there is no $\theta_1 \cap \theta_2$ and $\theta_1 \cap \theta_2 \cap \theta_3$.

Hence, the generation of $D_r^\Theta$ can run very fast if the basis is small, \emph{i.e.} if there is some constraints. The Matlab code is given in section~\ref{codes}.

\subsection{Decoding}
\label{decoding}
Once the decision on one element $A$ of $D_r^\Theta$ is taken, we have to transmit this decision to the human operator. Hence we must to decode the element $A$ (given by the integer numbers of the codification) in terms of unions and intersections of elements of $\Theta$. If we know that $A$ is in a subset of elements of $D_r^\Theta$ given by the operator, we only have to scan this subset. Now, if the decision $A$ comes from the focal elements (\textit{a priori} unknown) or from all the elements of $D_r^\Theta$ we must scan all $D_r^\Theta$ with possibly high complexity. What we propose here is to consider the elements of $D_r^\Theta$ ordering with first the elements most encountered in applications. Hence, we first scan the elements of $2^\Theta$ and in the same time the intersection basis that we must build for the generation of $D_r^\Theta$. Then, only if the element is not found we generate $D_r^\Theta$ and stop the generation when found (see the section~\ref{codes} for more details). 

The Smarandache's codification is an alternative to the decoding because user can directly understand it. Hence we can represent the focal element as an union of the distinct part of the Venn diagram. The Smarandache's codification allows a clear understanding of the different parts of the Venn diagram on the contrary than the proposed codification. This representation of the results (for the combination or the decision) does not need the generation of $D_r^\Theta$. However, if we need to generate $D_r^\Theta$ according to the strategy of decision, the decoding will give a better display without more generation of $D_r^\Theta$.

%\section{Evaluation}

\section{Concluding remarks}
This chapter presents a general belief function framework based on a practical codification of the focal elements. First the codification of the elements of the Venn diagram gives a codification of $\Theta$. Then, the eventual constraints are integrated giving a reduced discernment space $\Theta_r$. From the space $\Theta_r$, we obtain the codification of the focal elements. Hence, we manipulate elements of a reduced hyper power set $D_r^\Theta$ and not the complete hyper power set $D^\Theta$, reducing the complexity according to the kind of given constraints. 

With the practical codification, the step of combination is easily made using union and intersection functions. 

The step of decision was particularly studied, because of the difficulties to decide on $D^\Theta$ or $D_r^\Theta$. An extension of the approach given in \cite{Appriou05} in order to give the possibility to decide on the unions in $2^\Theta$ was proposed. Another approach based on the specificity was proposed in order to simply choose the elements on which decide according to their specificity.  

The principal goal of this chapter is to provide practical codes of a general belief function framework for the researchers and users needing the belief function theory. However, for sake of clarity, all the Matlab codes are not in the listing, but can be provided on demand to the author. The proposed codes are not optimized either for Matlab, or in general and can still have bugs. All suggestions in order to improve them are welcome.

\section{Matlab codes}
\label{codes}
We give and explain here some Matlab codes of the general belief function framework\footnote{Copyright \copyright \, 2008 Arnaud Martin. May be used free of charge. Selling without prior written consent prohibited. Obtain permission before redistributing.}. Note that the proposed codes are not optimized either for Matlab, or in general. 

First the human operator have to describe the problem (see function~\ref{config}) giving the cardinality of $\Theta$, the list of the focal elements and the corresponding bba for each experts, the eventual constraints (`~' if there is no constraint), the list of elements on which he want to obtain a decision and the parameters corresponding to the choice of combination rule, the choice of decision criterium the mode of fusion (static or dynamic) and the display. When the description of the problem is made, he just has to call the fuse function~\ref{fuse}.

\bigskip
\hrule height 1 pt 
\vspace{-0.2cm}
\begin{functionmat}- Command configuration
  \label{config}
\vspace{0.05cm}
\hrule height 1 pt 
\begin{verbatim}
% description of the problem
CardTheta=4; % cardinality of Theta
% list of experts with focal elements and associated bba
    expert(1).focal={'1' '1u3' '3' '1u2u3'};     
    expert(1).bba=[0.5421 0.2953 0.0924 0.0702];

    expert(2).focal={'1' '2' '1u3' '1u2u3'};
    expert(2).bba=[0.2022 0.6891 0.0084 0.1003];

    expert(3).focal={'1' '3n4' '1u2u3'};
    expert(3).bba=[0.2022 0.6891 0.1087];

constraint={'1n2' '1n3' '2n3'}; % set of empty elements
elemDec={'F'}; % set of decision elements

%-------------------------------------------------------------
% parameters 
criteriumComb=1; % combination citerium
criteriumDec=0; % decision criterium
mode='static'; % mode of fusion
display=3; % kind of display

%-------------------------------------------------------------
% fusion

fuse(expert,constraint,CardTheta,criteriumComb,criteriumDec,...
		mode,elemDec,display)
\end{verbatim}
\hrule
\end{functionmat}

The first step of the fuse function~\ref{fuse} is the coding. The cardinality of $\Theta$ gives the codification of the singletons of $\Theta$, thanks to the function~\ref{codingTheta}, then we add the constraints to $\Theta$ with the function~\ref{addConstraint} and obtain $\Theta_r$. With $\Theta_r$, the function~\ref{codingExpert} calling the function~\ref{codingFocal} codes the focal elements of the experts given by the human operator. The combination is made by the function~\ref{combination} in static mode. For dynamic fusion, we just consider one expert with the previous combination. In this case the order of the experts given by the user can have an important signification. The decision step is made with the function~\ref{decision}. The last step concern the display and the hard problem of the decoding. Thus, 4 choices are possible: no display, the results of the combination only, the results of decision only and both results. These displays could take long time according to the parameters given by the human operator. Hence, the results of the combination could have the complexity of the generation of $D_r^\Theta$ and must be avoid if the user does not need it. The complexity of the decision results could also be high if the user does not give the exact set of elements on witch decide, or only the singletons with `S' or on $2^\Theta$ with `2T'. In other cases, with luck, the execution time can be short thanks to the function~\ref{decodingFocal}.

\bigskip
\hrule height 1 pt 
\vspace{-0.2cm}
\begin{functionmat}- Fuse function
\label{fuse}
\vspace{0.05cm}
\hrule height 1 pt 
\begin{verbatim}
function fuse(expert,constraint,n,criteriumComb,criteriumDec,mode,elemDec,display)

% To fuse experts' opinions
% 
% fuse(expert,constraint,n,criteriumComb,criteriumDec,mode,elemDec,display)
% 
% Inputs:
% expertC = containt the structure of the list of coded focal elements and
% corresponding bba for all the experts
% constraint = the empty elements 
% elemDec = list of elements on which we can decide
% n = size of the discernment space
% criteriumComb = is the combination criterium
%         criteriumComb=1 Smets criterium
%         criteriumComb=2 Dempster-Shafer criterium (normalized)
%         criteriumComb=3 Yager criterium
%         criteriumComb=4 disjunctive combination criterium
%         criteriumComb=5 Florea criterium
%         criteriumComb=6 PCR6 
%         criteriumComb=7 Mean of the bbas
%         criteriumComb=8 Dubois criterium (normalized and 
%                                  disjunctive combination)
%         criteriumComb=9 Dubois and Prade criterium (mixt combination)
%         criteriumComb=10 Mixt Combination (Martin and Osswald criterium)
%         criteriumComb=11 DPCR (Martin and Osswald criterium)
%         criteriumComb=12 MDPCR (Martin and Osswald criterium)
%         criteriumComb=13 Zhang's rule
%
%
% criteriumDec = is the combination criterium
%           criteriumDec=0 maximum of the bba 
%           criteriumDec=1 maximum of the pignistic probability 
%           criteriumDec=2 maximum of the credibility
%           criteriumDec=3 maximum of the credibility with reject
%           criteriumDec=4 maximum of the plausibility
%           criteriumDec=5 Appriou criterium
%           criteriumDec=6 DSmP criterium
%
% mode = 'static' or 'dynamic'
% elemDec = list of elements on which we can decide, 
%     or A for all, S for singletons only, F for focal elements only,
%     SF for singleton plus focal elements, Cm for given specificity, 
%     2T for only 2^Theta (DST case)
% display = kind of display
%       display = 0 for no display, 
%       display = 1 for combination display, 
%       display = 2 for decision display, 
%       display = 3 for both displays,
%       display = 4 for both displays with Smarandache codification
%
% Output:
% res = containt the structure of the list of focal elements and
% corresponding bbas for the combinated experts
%
% Copyright (c) 2008 Arnaud Martin 

% Coding
[Theta,Scod]=codingTheta(n);
ThetaRed=addConstraint(constraint,Theta);

expertCod=codingExpert(expert,ThetaRed);

%--------
switch nargin
    case 1:5
        mode='static';
        elemDec=ThetaRed;
        display=4;
    case 6
        elemDec=ThetaRed;
        display=4;
    case 7
        elemDec=string2code(elemDec);
        display=4;
end
%--------
if (display==1) || (display==2) || (display==3)
    [DThetar,D_n]=generationDThetar(ThetaRed);
else
    switch elemDec{1}
        case {'A'}
            [DThetar,D_n]=generationDThetar(ThetaRed);
        otherwise
            DThetar.s={[]};
            DThetar.c={[]};
    end
end
%--------

% Combination
if strcmp(mode, 'static')
    [expertComb]=combination(expertCod,ThetaRed,criteriumComb);
else % dynamic case
    nbexp=size(expertCod,2);
    expertTmp(1)=expertCod(1);
    for exp=2:nbexp
        expertTmp(2)=expertCod(exp);
        expertTmp(1)=combination(expertTmp,ThetaRed,criteriumComb);
    end
    expertComb=expertTmp(1);
end

% Decision
[decFocElem]=decision(expertComb,ThetaRed,DThetar.c,criteriumDec,elemDec);

% Display
switch display
    case 0
        'no display'
    case 1
        % Result of the combination
        sFocal=size(expertComb.focal,2);
        focalRec=decodingExpert(expertComb,ThetaRed,DThetar);
        focal=code2string(focalRec)
        for i=1:sFocal
            disp ( [ focal{i},'=',num2str(expertComb.bba(i)) ] )
        end
    case 2
        % Result of the decision
        if isstruct(decFocElem)
            focalDec=decodingFocal(decFocElem.focal,elemDec,ThetaRed);
            disp(['decision:',code2string(focalDec)])
        else
            if decFocElem==0
                disp(['decision: rejected'])
            else
                if decFocElem==-1
                    disp(['decision: cannot be taken'])
                end
            end     
        end
    case 3
        % Result of the combination
        sFocal=size(expertComb.focal,2);
        expertDec=decodingExpert(expertComb,ThetaRed,DThetar);
        focal=code2string(expertDec.focal)
        for i=1:sFocal
            disp ( [ focal{i},'=',num2str(expertDec.bba(i)) ] )
        end
        % Result of the decision
        if isstruct(decFocElem)
            focalDec=decodingFocal(decFocElem.focal,elemDec,ThetaRed,DThetar);
            disp(['decision:',code2string(focalDec)])
        else
            if decFocElem==0
                disp(['decision: rejected'])
            else
                if decFocElem==-1
                    disp(['decision: cannot be taken'])
                end
            end     
        end
    case 4
        % Results with Smarandache codification display
        % Result of the combination
        sFocal=size(expertComb.focal,2);
        expertDec=cod2ScodExpert(expertComb,Scod);
        for i=1:sFocal
            disp ([expertDec.focal{i},'=',num2str(expertDec.bba(i))])
        end
        % Result of the decision
         if isstruct(decFocElem)
            focalDec=cod2ScodFocal(decFocElem.focal,Scod);
            disp(['decision:',focalDec])
        else
            if decFocElem==0
                disp(['decision: rejected'])
            else
                if decFocElem==-1
                    disp(['decision: cannot be taken'])
                end
            end     
        end
        
    otherwise
        'Accident in fuse: choice of display is uncorrect'
end
\end{verbatim}
\hrule
\end{functionmat}

\subsection{Codification}

The codification is based on the function~\ref{codingTheta}. The order of the integer numbers could be different, here the choice is made to number the intersection of all the elements with 1 and the smallest integer among the $|\Theta|=n$ bigger integers for the first singleton. In the same time this function give the correspondence between the integer numbers of the practical codification and the Smarandache's codification. This function~\ref{codingTheta} is based on the Matlab function \emph{nchoosek(tab,k)} given the array of all the combination of $k$ elements of the vector \emph{tab}. If the length of \emph{tab} is $n$, this function return an array of $C_n^k$ rows and $k$ columns.
\bigskip
\hrule height 1 pt 
\vspace{-0.2cm}
\begin{functionmat}- codingTheta function
\label{codingTheta}
\vspace{0.05cm}
\hrule height 1 pt 
\begin{verbatim}
function [Theta,Scod]=codingTheta(n)

% Code Theta for DSmT framework
%
% [Theta,Scod]=codingTheta(n)
%
% Input:
% n = cardinality of Theta
%
% Outputs:
% Theta = the liste of coded elements in Theta
% Scod = the bijection function between the integer of 
%   the coded elements in Theta and the Smarandache codification
%
% Copyright (c) 2008 Arnaud Martin 

i=2^n-1;
tabInd=[];
for j=n:-1:1
    tabInd=[tabInd j];
    Theta{j}=[i];
    
    Scod{i}=[j];
    i=i-1;
end

i=i+1;
for card=2:n
    tabPerm=nchoosek(tabInd,card);
    for j=1:n
        [l,c]=find(tabPerm==j);
        tabi=i.*ones(1,size(l,1));
        Theta{j}=[sort(tabi-l') Theta{j}];
        for nb=1:size(l,1)
            Scod{i-l(nb)}=[Scod{i-l(nb)} j];
        end
    end  
    i=i-size(tabPerm,1);
end
\end{verbatim}
\hrule
\end{functionmat}

The addition of the constraints is made in two steps: first the codification of the elements in the list \emph{constraint} is made with the function~\ref{codingFocal}, then the integer numbers in the codification of the constraints are suppressed from the codification of $\Theta$. The function \emph{string2code} is just the translation of the brackets and union and intersection operators in negative numbers (-3 for `(', -4 for `)', -1 for `$\cup$' and -2 for `$\cap$') in order to manipulate faster integers than strings. This simple function is not provided here.
\bigskip
\hrule height 1 pt 
\vspace{-0.2cm}
\begin{functionmat}- addConstraint function
\label{addConstraint}
\vspace{0.05cm}
\hrule height 1 pt 
\begin{verbatim}
function [ThetaR]=addConstraint(constraint,Theta)

% Code ThetaR the reduced form of Theta 
% taking into account the constraints given by the user
%
% [ThetaR]=addConstraint(constraint,Theta)
% 
% Inputs:
% constraint = the list of element considered as constraint 
%             or '2T' to work on 2^Theta
% Theta = the description of Theta after coding
%
% Output:
% ThetaR = the description of coded Theta after reduction 
% taking into account the constraints
%
% Copyright (c) 2008 Arnaud Martin 

if strcmp(constraint{1}, '2T')
    n=size(Theta,2);
    nbCons=1;
    for i=1:n
        for j=i+1:n
            constraint(nbCons)={[i -2 j]};
            nbCons=nbCons+1;
        end
    end
else
    constraint=string2code(constraint);
end

constraintC=codingFocal(constraint,Theta);

sConstraint=size(constraintC,2);
unionCons=[];
for i=1:sConstraint
    unionCons=union(unionCons,constraintC{i});
end

sTheta=size(Theta,2);
for i=1:sTheta
    ThetaR{i}=setdiff(Theta{i},unionCons);
end
\end{verbatim}
\hrule
\end{functionmat}

The function~\ref{codingFocal} simply transforms the list of focal elements given by the user with the codification of $\Theta$ to obtain the list of constraints and with $\Theta_r$ for the focal elements of each expert. The function~\ref{codingExpert} prepares the coding of focal elements and return the list of the experts with the coded focal elements. 
\bigskip
\hrule height 1 pt 
\vspace{-0.2cm}
\begin{functionmat}- codingFocal function
\label{codingFocal}
\vspace{0.05cm}
\hrule height 1 pt 
\begin{verbatim}
function [focalC]=codingFocal(focal,Theta)

% Code the focal element for DSmT framework
% 
% [focalC]=codingFocal(focal,Theta)
% 
% Inputs:
% focal = the list of focal element for one expert
% Theta = the description of Theta after coding
%
% Output:
% focalC = the list of coded focal element for one expert
%
% Copyright (c) 2008 Arnaud Martin 

 nbfoc=size(focal,2);
 if nbfoc
     for foc=1:nbfoc
         elemC=treat(focal{foc},Theta);
         focalC{foc}=elemC;
     end
 else
     focalC={[]};
 end 
end

%%
function [elemE]=eval(oper,a,b)

    if oper==-2
        elemE=intersect(a,b);
    else
        elemE=union(a,b);
    end
end

%%
function [elemC,cmp]=treat(focal,Theta)

    nbelem=size(focal,2);
    PelemC=0;
    oper=0;
    e=1;
    if nbelem
        while e <= nbelem
           elem=focal(e);
           switch elem
               case -1
                   oper=-1;
               case -2
                   oper=-2;
               case -3
                  [elemC,nbe]=treat(focal(e+1:end),Theta);
                  e=e+nbe;
                  
                   if oper~=0 & ~isequal(PelemC,0)
                       elemC=eval(oper,PelemC,elemC);               
                       oper=0;
                   end
                   PelemC=elemC;

               case -4
                   cmp=e;
                   e=nbelem;
               otherwise
                   elemC=Theta{elem};   
                   if oper~=0 & ~isequal(PelemC,0)
                       elemC=eval(oper,PelemC,elemC);               
                       oper=0;
                   end
                   PelemC=elemC;
           end
           e=e+1;
        end
    else
        elemC=[];
    end
end
\end{verbatim}
\hrule
\end{functionmat}

\bigskip
\hrule height 1 pt 
\vspace{-0.2cm}
\begin{functionmat}- codingExpert function
\label{codingExpert}
\vspace{0.05cm}
\hrule height 1 pt 
\begin{verbatim}
function [expertC]=codingExpert(expert,Theta)

% Code the focal element for DSmT framework
% 
% [expertC]=codingExpert(expert,Theta)
% 
% Inputs:
% expert = structure containing the list of focal elements for 
%         each expert and the bba corresponding
% Theta = the description of Theta after coding
% 
% Output:
% expertC = structure containing the list of coded focal element 
%         for each expert and the bba corresponding
% 
% Copyright (c) 2008 Arnaud Martin 

    nbExp=size(expert,2);
    for exp=1:nbExp
       focal=string2code(expert(exp).focal);
       expertC(exp).focal=codingFocal(focal,Theta);
       expertC(exp).bba=expert(exp).bba;
    end
end
\end{verbatim}
 \hrule
\end{functionmat}

\subsection{Combination}

The function~\ref{combination} proposes many combination rules. Most of them are based on the function~\ref{conjunctivefunc}, but for some combination rules we need to keep more information, so we use the function~\ref{globalConjunctive} for the conjunctive combination. \emph{E.g.} in the function~\ref{PCR6} note the simplicity of the code for the $\PCRmo$ combination rule. Other combination rules' codes are not given here for the sake of clarity.
\bigskip
\hrule height 1 pt 
\vspace{-0.2cm}
\begin{functionmat}- combination function
\label{combination}
\vspace{0.05cm}
\hrule height 1 pt
\begin{verbatim}
function [res]=combination(expertC,ThetaR,criterium)

% Give the combination of many experts
% 
% [res]=combination(expert,constraint,n,criterium)
% 
% Inputs:
% expertC = containt the structure of the list of focal elements 
%           and corresponding bba for all the experts
% ThetaR = the coded and reduced discernment space
% criterium = is the combination criterium
%    criterium=1 Smets criterium (conjunctive rule in open world)
%    criterium=2 Dempster-Shafer criterium (normalized) 
%                (conjunctive rule in closed world)
%    criterium=3 Yager criterium
%    criterium=4 disjunctive combination criterium
%    criterium=5 Florea criterium
%    criterium=6 PCR6 
%    criterium=7 Mean of the bbas
%    criterium=8 Dubois criterium 
%                (normalized and disjunctive combination)
%    criterium=9 Dubois and Prade criterium (mixt combination)
%    criterium=10 Mixt Combination (Martin and Osswald criterium)
%    criterium=11 DPCR (Martin and Osswald criterium)
%    criterium=12 MDPCR (Martin and Osswald criterium)
%    criterium=13 Zhang's rule
%
% Output:
% res = containt the structure of the list of focal elements and
%       corresponding bbas for the combinated experts
%
% Copyright (c) 2008 Arnaud Martin 

switch criterium
    case 1
        %Smets criterium
        res=conjunctive(expertC);
    case 2
        %Dempster-Shafer criterium (normalized)
        expConj=conjunctive(expertC);
        ind=findeqcell(expConj.focal,[]);
        if ~isempty(ind)
            k=expConj.bba(ind);
            expConj.bba=expConj.bba/(1-k);
            expConj.bba(ind)=0;
        end
        res=expConj;
    case 3
        %Yager criterium
        expConj=conjunctive(expertC);
        ind=findeqcell(expConj.focal,[]);
        if ~isempty(ind)
            k=expConj.bba(ind);
            eTheta=ThetaR{1};
            for i=2:n
                eTheta=[union(eTheta,ThetaR{i})];
            end
            indTheta=findeqcell(expConj.focal,eTheta);
            if ~isempty(indTheta)
                expConj.bba(indTheta)=expConj.bba(indTheta)+k;
                expConj.bba(ind)=0;
            else
                sFocal=size(expConj.focal,2);
                expConj.focal(sFocal+1)={eTheta};
                expConj.bba(sFocal+1)=k;
                expConj.bba(ind)=0;
            end
        end
        res=expConj;
    case 4
        %disjounctive criterium
        [res]=disjunctive(expertC);
    case 5
        % Florea criterium
        expConj=conjunctive(expertC);
        expDis=disjunctive(expertC);
        
        ind=findeqcell(expConj.focal,[]);
        if ~isempty(ind)
            k=expConj.bba(ind);
            alpha=k/(1-k+k*k);
            beta=(1-k)/(1-k+k*k); 
            
            expFlo=expConj;
            expFlo.bba=beta.*expFlo.bba;
            expFlo.bba(ind)=0;    
            nbFocConj=size(expConj.focal,2);
            nbFocDis=size(expDis.focal,2);
            
            expFlo.focal(nbFocConj+1:nbFocConj+nbFocDis)=expDis.focal;
            expFlo.bba(nbFocConj+1:nbFocConj+nbFocDis)=alpha.*expDis.bba;
            
            expFlo=reduceExpert(expFlo);
        else
            expFlo=expConj;
        end
        res=expFlo;
    case 6
        % PCR6
        [res]=PCR6(expertC);
    case 7
        % Means of the bba
        [res]=meanbba(expertC);
    case 8
        % Dubois criterium (normalized and disjunctive combination)
        expDis=disjunctive(expertC);
        
        ind=findeqcell(expDis.focal,[]);
        if ~isempty(ind)
            k=expDis.bba(ind);
            expDis.bba=expDis.bba/(1-k);
            expDis.bba(ind)=0;
        end
        res=expDis;
    case 9
        % Dubois and Prade criterium (mixt combination)
        [res]=DP(expertC);
    case 10
        % Martin and Ossawald criterium (mixt combination)
        [res]=Mix(expertC);
    case 11
        % DPCR (Martin and Osswald criterium)
        [res]=DPCR(expertC);
    case 12
        % MDPCR (Martin and Osswald criterium)
        [res]=MDPCR(expertC);
    case 13
        % Zhang's rule
        [res]=Zhang(expert)

    otherwise 
        'Accident: in combination choose of criterium: uncorrect'
end
\end{verbatim}
 \hrule
\end{functionmat}

\bigskip
\hrule height 1 pt 
\vspace{-0.2cm}
\begin{functionmat}- conjunctive function
\label{conjunctivefunc}
\vspace{0.05cm}
\hrule height 1 pt
\begin{verbatim}
function [res]=conjunctive(expert)

% Conjunctive Rule
% 
% [res]=conjunctive(expert)
% 
% Inputs:
% expert = containt the structures of the list of focal element and
% corresponding bba for all the experts
% 
% Output:
% res = is the resulting expert (structure of the list of focal 
%       element and corresponding bba)
%
% Copyright (c) 2008 Arnaud Martin 

nbexpert=size(expert,2);
for i=1:nbexpert
    nbfocal(i)=size(expert(i).focal,2);
    nbbba(i)=size(expert(i).bba,2);
    if nbfocal(i)~=nbbba(i)
        'Accident: in conj: the numbers of bba and focal element...
                are different'
    end
end

interm=expert(1);
for exp=2:nbexpert
    nbfocalInterm=size(interm.focal,2);
    i=1;
    comb.focal={};
    comb.bba=[];
    for foc1=1:nbfocalInterm
        for foc2=1:nbfocal(exp)
            tmp=intersect(interm.focal{foc1},expert(exp).focal{foc2});
            if isempty(tmp)
                tmp=[];
            end
            comb.focal(i)={tmp};
            comb.bba(i)=interm.bba(foc1)*expert(exp).bba(foc2);
            i=i+1;
        end
    end
    interm=reduceExpert(comb);
end
res=interm;
\end{verbatim}
 \hrule
\end{functionmat}

\bigskip
\hrule height 1 pt 
\vspace{-0.2cm}
\begin{functionmat}- globalConjunctive function
\label{globalConjunctive}
\vspace{0.05cm}
\hrule height 1 pt
\begin{verbatim}
function [res,tabInd]=globalConjunctive(expert)

% Conjunctive Rule conserving all the focal elements 
% during the combination
% 
% [res,tabInd]=globalConjunctive(expert)
% 
% Input:
% expert = containt the structures of the list of focal element and
% corresponding bba for all the experts
%
% outputs:
% res = is the resulting expert (structure of the list of focal 
%       element and corresponding bba)
% tabInd = table of the indices given the combination
%
% Copyright (c) 2008 Arnaud Martin 

nbexpert=size(expert,2);
for i=1:nbexpert
    nbfocal(i)=size(expert(i).focal,2);
    nbbba(i)=size(expert(i).bba,2);
    if nbfocal(i)~=nbbba(i)
        'Accident: in conj: the numbers of bba and focal element...
        					 are different'
    end
end
interm=expert(1);
tabIndPrev=[1:1:nbfocal(1)];   
for exp=2:nbexpert
    nbfocalInterm=size(interm.focal,2);
    i=1;
    comb.focal={};
    comb.bba=[];
    tabInd=[];
    for foc1=1:nbfocalInterm
        for foc2=1:nbfocal(exp)
            tmp=intersect(interm.focal{foc1},expert(exp).focal{foc2});
            tabInd=[tabInd [tabIndPrev(:,foc1);foc2]]; 
            if isempty(tmp)
                tmp=[];
            end
            comb.focal(i)={tmp};
            comb.bba(i)=interm.bba(foc1)*expert(exp).bba(foc2);
            i=i+1;
        end
    end
    tabIndPrev=tabInd;
    interm=comb;
end
res=interm;
\end{verbatim}
 \hrule
\end{functionmat}

\bigskip
\hrule height 1 pt 
\vspace{-0.2cm}
\begin{functionmat}- PCR6 function
\label{PCR6}
\vspace{0.05cm}
\hrule height 1 pt
\begin{verbatim}
function [res]=PCR6(expert)

% PCR6 combination rule
% 
% [res]=PCR6(expert)
% 
% Input:
% expert = containt the structures of the list of focal element and
% corresponding bba for all the experts
%
% Output:
% res = is the resulting expert (structure of the list of focal 
%      element and corresponding bba)
%
% Reference: A. Martin and C. Osswald, ''A new generalization of the
%  proportional conflict redistribution rule stable in terms of decision,'' 
%  Applications and Advances of DSmT for Information Fusion, Book 2,
%  American Research Press Rehoboth, F. Smarandache and J. Dezert, 
%  pp. 69-88 2006.
%
% Copyright (c) 2008 Arnaud Martin 

[expertConj,tabInd]=globalConjunctive(expert);

ind=findeqcell(expertConj.focal,[]);
nbexp=size(tabInd,1);
if ~isempty(ind)
    expertConj.bba(ind)=0;
    sInd=size(ind,2);
    for i=1:sInd
        P=1;
        S=0;
        for exp=1:nbexp
            bbaexp=expert(exp).bba(tabInd(exp,ind(i)));
            P=P*bbaexp;
            S=S+bbaexp;
        end
        for exp=1:nbexp 
            expertConj.focal(end+1)=expert(exp).focal(tabInd(exp,ind(i)));
            expertConj.bba(end+1)=expert(exp).bba(tabInd(exp,ind(i)))*P/S;
        end
    end
end
res=reduceExpert(expertConj);
\end{verbatim}
  \hrule
\end{functionmat}

\subsection{Decision}

The function~\ref{decision} gives the decision on the expert focal element list for the corresponding bba with one of the chosen criterium and on the elements given by the user for the decision. Note that the choices `A' and `Cm' for the variable \emph{elemDec} could take a long time because it need the generation of $D_r^\Theta$. This function can call one of the decision functions~\ref{pignistic}, \ref{credibility}, \ref{plausibility}, \ref{DSmPep}. If any decision is possible on the chosen elements given by \emph{elemDec}, the function return -1. In case of reject element, te function return 0.

\bigskip
\hrule height 1 pt 
\vspace{-0.2cm}
\begin{functionmat}- decision function
\label{decision}
\vspace{0.05cm}
\hrule height 1 pt
\begin{verbatim}
function [decFocElem]=decision(expert,Theta,criterium,elemDec)

% Give the decision for one expert
%
% [decFocElem]=decision(expert,Theta,criterium)
%
% Inputs:
% expert = containt the structure of the list of focal elements and
%         corresponding bba for all the experts
% Theta = list of coded (and reduced with constraint) of the 
%         elements of the discernement space
% criterium = is the combination criterium
%    criterium=0 maximum of the bba 
%    criterium=1 maximum of the pignistic probability 
%    criterium=2 maximum of the credibility
%    criterium=3 maximum of the credibility with reject
%    criterium=4 maximum of the plausibility
%    criterium=5 DSmP criterium 
%    criterium=6 Appriou criterium
%    criterium=7 Credibility on DTheta criterium
%    criterium=8  pignistic on DTheta criterium
% elemDec = list of elements on which we can decide, 
%    or A for all, S for singletons only, F for focal elements only,
%    SF for singleton plus focal elements, Cm for given specificity, 
%    2T for only 2^Theta (DST case)
%
% Output:
% decFocElem = the retained focal element, 0 in case of rejet, -1 
%              if the decision cannot be taken on elemDec
%
% Copyright (c) 2008 Arnaud Martin 


type=1;
switch elemDec{1}
    case 'S'
        type=0;
        elemDecC=Theta;
        expertDec=expert;
    case 'F'
        elemDecC=expert.focal;
        expertDec=expert;
    case 'SF'
        expertDec=expert;
        n=size(Theta,2);
        for i=1:n
            expertDec.focal{end+1}=Theta{i};
            expertDec.bba(end+1)=0;
        end
        expertDec=reduceExpert(expertDec);
        elemDecC=expertDec.focal;
    case 'Cm'
        sElem=size(elemDec,2);
        switch sElem
            case 2
                minSpe=str2num(elemDec{2});
                maxSpe=minSpe;
            case 3
                minSpe=str2num(elemDec{2});
                maxSpe=str2num(elemDec{3});
            otherwise
                'Accident in decision: with the option Cm for ....
                elemDec give the specifity of decision element ...
                (eventually the minimum and the maximum of the ...
                desired specificity'
                pause
        end
        
        elemDecC=findFocal(Theta,minSpe,maxSpe); 
        expertDec.focal=elemDecC;
        
        expertDec.bba=zeros(1,size(elemDecC,2));
        for foc=1:size(expert.focal,2)
            ind=findeqcell(elemDecC,expert.focal{foc});
            if ~isempty(ind)
                expertDec.bba(ind)=expert.bba(foc);
            else
                expertDec.bba(ind)=0;
            end
        end
        
    case '2T'
        type=0;
        natoms=size(Theta,2);
        expertDec.focal(1)={[]};
        indFoc=findeqcell(expert.focal,{[]});
        if isempty(indFoc)
            expertDec.bba(1)=0;
        else
            expertDec.bba(1)=expert.bba(indFoc);
        end
        step =2;
        for i=1:natoms
            expertDec.focal(step)=codingFocal({[i]},Theta);
            
            indFoc=findeqcell(expert.focal,expertDec.focal{step});
            if isempty(indFoc)
                expertDec.bba(step)=0;
            else
                expertDec.bba(step)=expert.bba(indFoc);
            end
            
            step=step+1;
            indatom=step;
            for step2=2:indatom-2
                expertDec.focal(step)={[union(expertDec.focal{step2},...
                                       expertDec.focal{indatom-1})]};
                
                indFoc=findeqcell(expert.focal,expertDec.focal{step});
                if isempty(indFoc)
                    expertDec.bba(step)=0;
                else
                    expertDec.bba(step)=expert.bba(indFoc);
                end
            
                step=step+1;
            end
        end
        elemDecC=expertDec.focal;
    case 'A'
        elemDecC=generationDThetar(Theta); 
        elemDecC=reduceFocal(elemDecC);
        expertDec.focal=elemDecC;
        expertDec.bba=zeros(1,size(elemDecC,2));
        for foc=1:size(expert.focal,2)
           expertDec.bba(findeqcell(elemDecC,expert.focal{foc}))...
                     =expert.bba(foc);
        end
    otherwise
        type=0;
        elemDec=string2code(elemDec);
        elemDecC=codingFocal(elemDec,Theta);
        expertDec=expert;
        nbElemDec=size(elemDecC,2);
        for foc=1:nbElemDec
            if ~isElem(elemDecC{foc}, expertDec.focal) 
                expertDec.focal{end+1}=elemDecC{foc};
                expertDec.bba(end+1)=0;
            end
        end
end

%---------------------------------------------------------
nbFocal=size(expertDec.focal,2);
switch criterium

    case 0
        % maximum of the bba
        nbFocal=size(expertDec.focal,2);
        nbElem=0;
        for foc=1:nbFocal
            ind=findeqcell(elemDecC,expertDec.focal{foc});
            if ~isempty(ind)
                bba(ind)=expertDec.bba(foc);
            end
        end
        [bbaMax,indMax]=max(bba);
        if bbaMax~=0
            decFocElem.bba=bbaMax;
            decFocElem.focal={elemDecC{indMax}};
        else
            decFocElem=-1;
        end
    case 1
        % maximum of the pignistic probability 
        [BetP]=pignistic(expertDec);
        decFocElem=MaxFoc(BetP,elemDecC,type);
    case 2
        % maximum of the credibility
        [Bel]=credibility(expertDec);
        decFocElem=MaxFoc(Bel,elemDecC,type);
    case 3
        % maximum of the credibility with reject
        [Bel]=credibility(expertDec);
        
        TabSing=[];
        focTheta=[];
        for i=1:size(Theta,2)
            focTheta=union(focTheta,Theta{i});
        end
        
        for foc=1:nbFocal
            if isElem(Bel.focal{foc}, elemDecC) 
               TabSing=[TabSing [foc ;  Bel.Bel(foc)]];
            end
        end 
        [BelMax,indMax]=max(TabSing(2,:));
        if BelMax~=0
            focMax=Bel.focal{TabSing(1,indMax)};

            focComplementary=setdiff(focTheta,focMax);
            if isempty(focComplementary)
                focComplementary=[];
            end

            ind=findeqcell(Bel.focal,focComplementary);
            if BelMax < Bel.Bel(ind) 
            % if ind is empty this is always false    
                decFocElem=0; % That means that we reject
            else
               if isempty(ind)
                   decFocElem=0; % That means that we reject
               else
                decFocElem.focal={Bel.focal{TabSing(1,indMax)}};
                decFocElem.Bel=BelMax;
               end
            end
        else
            decFocElem=-1; % That means that we reject
        end
    case 4
        % maximum of the plausibility
        [Pl]=plausibility(expertDec);
        decFocElem=MaxFoc(Pl,elemDecC,type);
    case 5
        % DSmP criterium
        epsilon=0.00001; % 0 can allows problem
        
        [DSmP]=DSmPep(expertDec,epsilon);
        decFocElem=MaxFoc(DSmP,elemDecC,type);
    case 6
        % Appriou criterium
        [Pl]=plausibility(expertDec); 
        lambda=1;
        r=0.5;
        bm=BayesianMass(expertDec,lambda,r); 
        Newbba=Pl.Pl.*bm.bba;
        % normalization
        Newbba=Newbba/sum(Newbba);
        funcDec.focal=Pl.focal;
        funcDec.bba=Newbba;
        decFocElem=MaxFoc(funcDec,elemDecC,type);
    case 7
        % Credibility on DTheta criterium
        [Bel]=credibility(expertDec); 
        lambda=1;
        r=0.5;
        bm=BayesianMass(expertDec,lambda,r); 
        Newbba=Bel.Bel.*bm.bba;
        % normalization
        Newbba=Newbba/sum(Newbba);
        funcDec.focal=Bel.focal;
        funcDec.bba=Newbba;
        decFocElem=MaxFoc(funcDec,elemDecC,type);
    case 8
        % pignistic on DTheta criterium
        [BetP]=pignistic(expertDec); 
        lambda=1;
        r=0.5;
        bm=BayesianMass(expertDec,lambda,r); 
        Newbba=BetP.BetP.*bm.bba;
        % normalization
        Newbba=Newbba/sum(Newbba);
        funcDec.focal=BetP.focal;
        funcDec.bba=Newbba;
        decFocElem=MaxFoc(funcDec,elemDecC,type);
    otherwise 
        'Accident: in decision choose of criterium: uncorrect'
end
end


%%
function [bool]=isElem(focal, listFocal)

% The g oal of this function is to return a boolean on the test focal in
% listFocal
% 
% [bool]=isElem(focal, listFocal)
% 
% Inputs:
% focal = one focal element (matrix)
% listFocal = the list of elements in Theta (all different)
%
% Output:
% bool = boolean: true if focal is in listFocal, elsewhere false
%
% Copyright (c) 2008 Arnaud Martin 

n=size(listFocal,2);
bool=false;
for i=1:n
    if isequal(listFocal{i},focal)
        bool=true;
        break;
    end
end
end

%%
function [decFocElem]=MaxFoc(funcDec,elemDecC,type)

fieldN=fieldnames(funcDec);

switch fieldN{2}
    case 'BetP'
        funcDec.bba=funcDec.BetP;
    case 'Bel'
        funcDec.bba=funcDec.Bel;
    case 'Pl'
        funcDec.bba=funcDec.Pl;
    case 'DSmP'
        funcDec.bba=funcDec.DSmP;
        
end

if type 
    [funcMax,indMax]=max(funcDec.bba);
    FocMax={funcDec.focal{indMax}};
else
    nbFocal=size(funcDec.focal,2);
    TabSing=[];
    for foc=1:nbFocal
        if isElem(funcDec.focal{foc}, elemDecC) 
            TabSing=[TabSing [foc ;  funcDec.bba(foc)]];
        end
    end 
    [funcMax,indMax]=max(TabSing(2,:));
    FocMax={funcDec.focal{TabSing(1,indMax)}};
end

if funcMax~=0
    decFocElem.focal=FocMax;
    switch fieldN{2}
        case 'BetP'
            decFocElem.BetP=funcMax;
        case 'Bel'
            decFocElem.Bel=funcMax;
        case 'Pl'
            decFocElem.Pl=funcMax;
        case 'DSmP'
            decFocElem.DSmP=funcMax;
    end
else
    decFocElem=-1;
end
    
end
\end{verbatim}
  \hrule
\end{functionmat}

\bigskip
\hrule height 1 pt 
\vspace{-0.2cm}
\begin{functionmat}- findFocal function
\label{findFocal}
\vspace{0.05cm}
\hrule height 1 pt
\begin{verbatim}
function [elemDecC]=findFocal(Theta,minSpe,maxSpe)

% Find the element of DTheta with the minium of specifity minSpe 
% and the maximum maxSpe
%
% [elemDecC]=findFocal(Theta,minSpe,maxSpe)
%
% Input:
% Theta = list of coded (and eventually reduced with constraint) of 
%         the elements of the discernment space
% minSpe = minimum of the wanted specificity
% minSpe = maximum of the wanted specificity
%
% Output:
% elemDec = list of elements on which we want to decide with the 
%           minimum of specifity minSpe and the maximum maxSpe
%
% Copyright (c) 2008 Arnaud Martin 

elemDecC{1}=[]; 

n=size(Theta,2);
ThetaSet=[];
for i=1:n
    ThetaSet=union(ThetaSet,Theta{i});
end
for s=minSpe:maxSpe
    tabs=nchoosek(ThetaSet,s);
    elemDecC(end+1:end+size(tabs,1))=num2cell(tabs,2)';
end    

elemDecC=elemDecC(2:end);
\end{verbatim}
  \hrule
\end{functionmat}

\bigskip
\hrule height 1 pt 
\vspace{-0.2cm}
\begin{functionmat}- pignistic function
\label{pignistic}
\vspace{0.05cm}
\hrule height 1 pt
\begin{verbatim}
function [BetP]=pignistic(expert)

% Generalized Pignistic Transformation
%
% [BetP]=pignistic(expert)
%
% Input:
% expert = containt the structures of the list of focal element and
%          corresponding bba for all the experts
% expert.focal = list of focal elements
% expert.bba = matrix of bba
% 
% Output:
% BetP = containt the structure of the list of focal element and 
%        the matrix of the plausibility corresponding
% BetP.focal = list of focal elements
% BetP.BetP = matrix of the pignistic transformation

% Comment : 1- the code of the focal elements must inculde 
%                 the constraints
%           2- The pignistic is given only on the elements 
%                  in the list of focal of expert (the
%                  bba can be 0)
%
% Copyright (c) 2008 Arnaud Martin 

nbFocal=size(expert.focal,2);

BetP.focal=expert.focal;
BetP.BetP=zeros(1,nbFocal);

for focA=1:nbFocal   
    for focB=1:nbFocal
       focI=intersect(expert.focal{focA},expert.focal{focB});
       if ~isempty(focI)
          BetP.BetP(focA)=BetP.BetP(focA)+size(focI,2)/...
          size(expert.focal{focB},2)*expert.bba(focB);
       else 
           if isequal(expert.focal{focB},[]) 
           % for the empty set:
           % cardinality(empty set)/cardinality(empty set)=1, 
           % so we add the bba
               BetP.BetP(focA)=BetP.BetP(focA)+expert.bba(focB);
           end
       end
    end
end
\end{verbatim}
  \hrule
\end{functionmat}

\bigskip
\hrule height 1 pt 
\vspace{-0.2cm}
\begin{functionmat}- credibility function
\label{credibility}
\vspace{0.05cm}
\hrule height 1 pt
\begin{verbatim}
function [Bel]=credibility(expert)

% Credibility function
%
% [Bel]=credibility(expert)
%
% Input:
% expert = containt the structures of the list of focal element and
%          corresponding bba for all the experts
% expert.focal = list of focal elements
% expert.bba = matrix of bba
%
% Output:
% Bel = containt the structure of the list of focal element and 
%       the matrix of the credibility corresponding
% Bel.focal = list of focal elements
% Bel.Bel = matrix of the credibility 

% Comment : 1- the code of the focal elements must inculde 
%              the constraints
%           2- The credibility is given only on the elements 
%              in the list of focal of expert (the
%              bba can be 0)
%
% Copyright (c) 2008 Arnaud Martin 

nbFocal=size(expert.focal,2);

Bel.focal=expert.focal;
Bel.Bel=zeros(1,nbFocal);

for focA=1:nbFocal   
    for focB=1:nbFocal
        indMem=ismember(expert.focal{focB},expert.focal{focA});
        
       if sum(indMem)==size(expert.focal{focB},2)
          Bel.Bel(focA)=Bel.Bel(focA)+expert.bba(focB);
       else 
           if isequal(expert.focal{focB},[]) 
           % the empty set is include to all the focal elements
               Bel.Bel(focA)=Bel.Bel(focA)+expert.bba(focB);
           end
       end
    end
end
\end{verbatim}
  \hrule
\end{functionmat}

\bigskip
\hrule height 1 pt 
\vspace{-0.2cm}
\begin{functionmat}- plausibility function
\label{plausibility}
\vspace{0.05cm}
\hrule height 1 pt
\begin{verbatim}
function [Pl]=plausibility(expert)

% Plausibility function
%
% [Pl]=plausibility(expert)
%
% Input:
% expert = containt the structures of the list of focal element and
%          corresponding bba for all the experts
% expert.focal = list of focal elements
% expert.bba = matrix of bba
%
% Output:
% Pl = containt the structure of the list of focal element and 
%      the matrix of the plausibility corresponding
% Pl.focal = list of focal elements
% Pl.Pl = matrix of the plausibility 

% Comment : 1- the code of the focal elements must inculde 
%              the constraints
%           2- The plausibility is given only on the elements 
%              in the list of focal of expert (the
%              bba can be 0)
%
% Copyright (c) 2008 Arnaud Martin 


nbFocal=size(expert.focal,2);
Pl.focal=expert.focal;
Pl.Pl=zeros(1,nbFocal);

for focA=1:nbFocal   
    for focB=1:nbFocal
       focI=intersect(expert.focal{focA},expert.focal{focB});
       if ~isempty(focI)
          Pl.Pl(focA)=Pl.Pl(focA)+expert.bba(focB);
       else 
           if isequal(expert.focal{focB},[])...
            && isequal(expert.focal{focA},[]) 
           % for the empty set we keep the bba for the Pl
               Pl.Pl(focA)=Pl.Pl(focA)+expert.bba(focB);
           end
       end
    end
end
\end{verbatim}
   \hrule
\end{functionmat}

\bigskip
\hrule height 1 pt 
\vspace{-0.2cm}
\begin{functionmat}- DSmPep function
\label{DSmPep}
\vspace{0.05cm}
\hrule height 1 pt
\begin{verbatim}
function [DSmP]=DSmPep(expert,epsilon)

% DSmP Transformation
%
% [DSmP]=DSmPep(expert,epsilon)
%
% Inputs:
% expert = containt the structures of the list of focal element and
%          corresponding bba for all the experts
% expert.focal = list of focal elements
% expert.bba = matrix of bba
% epsilon = epsilon coefficient
% 
% Output:
% DSmPep = containt the structure of the list of focal element and 
%          the matrix of the plausibility corresponding
% DSmPep.focal = list of focal elements
% DSmPep.DSmP = matrix of the pignistic transformation
%
% Reference: Dezert & Smarandache, 
%   ''A new probbilistic transformation of belief mass assignment'', 
%   fusion 2008, Cologne, Germany.
%
% Copyright (c) 2008 Arnaud Martin 


nbFocal=size(expert.focal,2);

DSmP.focal=expert.focal;
DSmP.DSmP=zeros(1,nbFocal);

for focA=1:nbFocal 
    for focB=1:nbFocal
       focI=intersect(expert.focal{focA},expert.focal{focB});
       sumbbaFocB=0;
       sFocB=size(expert.focal{focB},2);
       for elB=1:sFocB
           ind=findeqcell(expert.focal,expert.focal{focB}(elB));
           if ~isempty(ind)
               sumbbaFocB=sumbbaFocB+expert.bba(ind);
           end
       end
       if ~isempty(focI) 
           sumbbaFocI=0;
           sFocI=size(focI,2);
           for elB=1:sFocI
               ind=findeqcell(expert.focal,focI(elB));
               if ~isempty(ind)
                   sumbbaFocI=sumbbaFocI+expert.bba(ind);
               end
           end
           DSmP.DSmP(focA)=DSmP.DSmP(focA)+expert.bba(focB)...
           *(sumbbaFocI+epsilon*sFocI)/(sumbbaFocB+epsilon*sFocB);
       end
    end
end
\end{verbatim}
  \hrule
\end{functionmat}

\subsection{Decoding and generation of $D_r^\Theta$}

For the displays, we must decode the focal elements and/or the final decision. The function~\ref{decodingExpert} decodes the focal elements in the structure expert that contain normally only one expert. This function calls the function~\ref{decodingFocal} that really does the decoding for the user. This function is based on the generation of $D_r^\Theta$ given by the function~\ref{generationDThetar} that a is modified and adapted code from \cite{Dezert04c}. To generate $D_r^\Theta$ we first must create the intersection basis. Hence in the function~\ref{decodingFocal} we use a loop of $2^\Theta$ in order to generate the basis and in the same time to scan the power set $2^\Theta$ and also the elements of the intersection basis. These two basis (intersection and union) are in fact concatenated during the construction, so we scan also some elements such intersections of previous unions and unions of previous intersections. This generated set of elements does not cover $D_r^\Theta$. When all the searching focal elements (that can be only one decision element) are found, we stop the function and avoid to generate all $D_r^\Theta$. Hence if the searching elements are not all found after this loop, we begin to generate $D_r^\Theta$ and stop when all elements are found. So, with luck, that can be fast.

We can avoid to generate $D_r^\Theta$ for only the display if we use the Smarandache's codification. The function~\ref{cod2ScodExpert} transforms the used code of the focal elements in the structure expert in the Smarandache's code, easer to understand by reading. This function calls the function~\ref{cod2ScodFocal} that really does the transformation. The focal elements are directly in string for the display. 

\bigskip
\hrule height 1 pt 
\vspace{-0.2cm}
\begin{functionmat}- decodingExpert function
\label{decodingExpert}
\vspace{0.05cm}
\hrule height 1 pt
\begin{verbatim}
function [expertDecod]=decodingExpert(expert,Theta,DTheta)

% The goal of this function is to decode the focal elements in expert
% 
% [expertDecod]=decodingExpert(expert,Theta)
% 
% Inputs:
% expert = containt the structure of the list of focal elements after 
% combination and corresponding bba for all the experts (generally use 
% for only one after combination)
% Theta = list of coded (and reduced with constraint) of the elements of
%         the discernement space
% DTheta = list of coded (and reduced with constraint) of the elements of
%           DTheta
%         
% Output:
% expertDecod = containt the structure of the list of decoded (for human) 
%       focal elements and corresponding bba for all the experts
%
% Copyright (c) 2008 Arnaud Martin 

    nbExp=size(expert,2);
    for exp=1:nbExp
        focal=expert(exp).focal;
        expertDecod(exp).focal=decodingFocal(focal,{'A'},Theta,DTheta);
        expertDecod(exp).bba=expert(exp).bba;
    end

end
\end{verbatim}
  \hrule
\end{functionmat}

\bigskip
\hrule height 1 pt 
\vspace{-0.2cm}
\begin{functionmat}- decodingFocal function
\label{decodingFocal}
\vspace{0.05cm}
\hrule height 1 pt
\begin{verbatim}
function [focalDecod]=decodingFocal(focal,elemDec,Theta,DTheta)

% The goal of this function is to decode the focal elements
% 
% [focalDecod]=decodingFocal(focal,elemDec,Theta)
% 
% Inputs:
% expert = containt the structure of the list of focal elements after 
% combination and corresponding bba for all the experts
% elemDec = the description of the subset of uncoded elements 
%           for decision
% Theta = list of coded (and reduced with constraint) of the 
%         elements of the discernement space
% DTheta = list of coded (and reduced with constraint) of the 
%           elements of DTheta, eventually empty if not necessary
% Output:
% focalDecod = containt the list of decoded (for human) focal elements 
%
% Copyright (c) 2008 Arnaud Martin 

switch elemDec{1}
    case {'F','A','SF','Cm'}
        opt=1;
    case 'S'
        opt=0;
        elemDecC=Theta;
        for i=1:size(Theta,2)
            elemDec(i)={[i]};
        end
    case '2T'
        opt=0;
        natoms=size(Theta,2);
        elemDecC(1)={[]};
        elemDec(1)={[]};
        step =2;
        for i=1:natoms
            elemDecC(step)=codingFocal({[i]},Theta);
            elemDec(step)={[i]};
            step=step+1;
            indatom=step;
            for step2=2:indatom-2
                elemDec(step)={[elemDec{step2} -1 elemDec{indatom-1}]};
                elemDecC(step)={[union(elemDecC{step2},elemDecC{indatom-1})]};
                step=step+1;
            end
        end

    otherwise
        opt=0;
        elemDecN=string2code(elemDec);
        elemDecC=codingFocal(elemDecN,Theta);
end

if ~opt
    sFoc=size(focal,2);
    for foc=1:sFoc
        [ind]=findeqcell(elemDecC,focal{foc});
        if isempty(ind)
            'Accident in decodingFocal: elemDec does not be 2T'
            pause
        else
            focalDecod(foc)=elemDec(ind);  
        end
    end
else
    focalDecod=cell(size(focal));
    cmp=0;
    sFocal=size(focal,2);
    sDTheta=size(DTheta.c,2);
    i=1;
    while i<sDTheta && cmp<sFocal
       DThetai=DTheta.c{i};
       indeq=findeqcell(focal,DThetai);
       if ~isempty(indeq)
           cmp=cmp+1;
           focalDecod(indeq)=DTheta.s(i);
       end 
       i=i+1;
    end
end
\end{verbatim}
  \hrule
\end{functionmat}

\bigskip
\hrule height 1 pt 
\vspace{-0.2cm}
\begin{functionmat}- cod2ScodExpert function
\label{cod2ScodExpert}
\vspace{0.05cm}
\hrule height 1 pt
\begin{verbatim}
function [expertDecod]=cod2ScodExpert(expert,Scod)

% The goal of this function is to code the focal elements in 
% expert with the Smarandache's codification from the practical 
% codification in order to display the expert
%
% [expertDecod]=cod2ScodExpert(expert,Scod)
% 
% Inputs:
% expert = containt the structure of the list of focal elements after 
% combination and corresponding bba for all the experts (generally use 
% for only one after combination)
% Scod = list of distinct part of the Venn diagram coded with the
%        Smarandache's codification 
% Output:
% expertDecod = containt the structure of the list of decoded (for human) 
%               focal elements and corresponding bba for all the experts
%
% Copyright (c) 2008 Arnaud Martin 

    nbExp=size(expert,2);
    for exp=1:nbExp
        focal=expert(exp).focal;
        expertDecod(exp).focal=cod2ScodFocal(focal,Scod);
        expertDecod(exp).bba=expert(exp).bba;
    end

end
\end{verbatim}
  \hrule
\end{functionmat}

\bigskip
\hrule height 1 pt 
\vspace{-0.2cm}
\begin{functionmat}- cod2ScodFocal function
\label{cod2ScodFocal}
\vspace{0.05cm}
\hrule height 1 pt
\begin{verbatim}
function [focalDecod]=cod2ScodFocal(focal,Scod)

% The goal of this function is to code the focal elements with the
% Smarandache's codification from the practical codification in order to
% display the focal elements
% 
% [focalDecod]=cod2ScodFocal(focal,Scod)
% 
% Inputs:
% expert = containt the structure of the list of focal elements after 
%     combination and corresponding bba for all the experts
% Scod = list of distinct part of the Venn diagram coded with the
%        Smarandache's codification 
% Output:
% focalDecod = containt the list of decoded (for human) focal elements 
%
% Copyright (c) 2008 Arnaud Martin 

sFocal=size(focal,2);
for foc=1:sFocal
    sElem=size(focal{foc},2);
    if sElem==0
        focalDecod{foc}='{}';
    else
        ch='{';
        ch=strcat(ch,'<');
        ch=strcat(ch,num2str(Scod{focal{foc}(1)}));
        ch=strcat(ch,'>');
        for elem=2:sElem
            ch=strcat(ch,',<');
            ch=strcat(ch,num2str(Scod{focal{foc}(elem)}));  
            ch=strcat(ch,'>');
        end
        focalDecod{foc}=strcat(ch,'}');
    end
end
\end{verbatim}
  \hrule
\end{functionmat}

\bigskip
\hrule height 1 pt 
\vspace{-0.2cm}
\begin{functionmat}- generationDThetar function
\label{generationDThetar}
\vspace{0.05cm}
\hrule height 1 pt
\begin{verbatim}
function [DTheta]=generationDThetar(Theta)

% Generation of DThetar: modified and adapted code from 
% Dezert & Smarandache Chapter 2 DSmT book % Vol 1 to generate DTeta
%
% [DTheta]=generationDThetar(Theta)
%
% Input:
% Theta = list of coded (and eventually reduced with constraint) of 
%         the elements of the discernment space
%
% Output:
% DTheta = list of coded (and eventually reduced with constraint in 
%          this case some elements can be the same) of the elements 
%          of the DTheta
%
% Copyright (c) 2008 Arnaud Martin 

n=size(Theta,2);
step =1;
for i=1:n
    basetmp(step)={[Theta{i}]};
    step=step+1;
    indatom=step;
    for step2=1:indatom-2
        basetmp(step)={intersect(basetmp{indatom-1},basetmp{step2})};
        step=step+1;
    end
end
sBaseTmp=size(basetmp,2);
step=1;
for i=1:sBaseTmp
    if ~isempty(basetmp{i})
        base(step)=basetmp(i);
        step=step+1;
    end
end
sBase=size(base,2);
DTheta{1}=[];
step=1;
nbC=2;
stop=0;
D_n1 =[0 ; 1];
sDn1=2;
for nn=1:n 
    D_n =[ ] ;

    cfirst=1+(nn==n);
    for i =1:sDn1 
        Li=D_n1(i,:);
        sLi=size(Li,2);
        if (2*sLi>sBase)&& (Li(sLi-(sBase-sLi))==1)
            stop=1;
            break
        end

        for j=i:sDn1
            Lj=D_n1(j,:); 
            if(and(Li,Lj)==Li)&(or(Li,Lj)==Lj)
                D_n=[D_n ; Li Lj ] ;
                
                if size(D_n,1)>step
                    step=step+1;
                    DTheta{step}=[];
                    for c=cfirst:nbC
                        if D_n(end,c)
                            if isempty(DTheta{step})
                                DTheta{step}=base{sBase+c-nbC};
                            else
                                DTheta{step}=union(DTheta{step},base{sBase+c-nbC});
                            end
                        end
                    end
                end
            end
        end
    end
    if stop
        break
    end
    D_n1=D_n;
    sDn1=size(D_n1,1);
    nbC=2*size(D_n1,2); 
end
\end{verbatim}
  \hrule
\end{functionmat}

\section*{Acknowledgment}
The author thanks Pascal Djiknavorian for the interesting discussion on the generation of $D_r^\Theta$, Florentin Smarandache for his comments on the codification and Jean Dezert for his advices on the representation of the DSm cardinality ${\cal C_M}(X)$. 

\bibliographystyle{plain} 
\bibliography{biblio}

\end{document}